%% file: main.tex
\documentclass{article} 
\usepackage{colm2024_conference}

\usepackage{microtype}
\usepackage{hyperref}
\usepackage{url}
\usepackage{booktabs}
\usepackage{wrapfig}
\usepackage{adjustbox}

\input{math_commands.tex}

\usepackage{microtype}
\usepackage{graphicx}
\usepackage{booktabs} 

\usepackage{hyperref}

\usepackage{amsmath}
\usepackage{amssymb}
\usepackage{mathtools}
\usepackage{amsthm}

\usepackage[capitalize,noabbrev]{cleveref}
\Crefname{equation}{Eq.}{Eqs.}

\theoremstyle{plain}

\theoremstyle{definition}

\theoremstyle{remark}

\usepackage{multicol}
\usepackage{multirow}
\usepackage{tabu}
\usepackage{tabularx}
\usepackage{makecell}
\usepackage{subcaption}
\usepackage{algorithm,algpseudocode}
\usepackage{bbm}

\newcommand\given[1][]{\:#1\vert\:}
\newcommand{\ours}{CGD}

\colmfinalcopy 

\title{Seeing is Believing: Mitigating Hallucination in Large Vision-Language Models via CLIP-Guided Decoding}

\author{Ailin Deng\textsuperscript{1}, Zhirui Chen\textsuperscript{2} \& Bryan Hooi\textsuperscript{1}\\
\textsuperscript{1}School of Computing, \textsuperscript{2}Department of Industrial Systems Engineering and Management\\
National University of Singapore\\
\texttt{\{ailin,zhiruichen\}@u.nus.edu,bhooi@comp.nus.edu.sg}
}

\begin{document}

\maketitle


\begin{abstract}

Large Vision-Language Models (LVLMs) are susceptible to object hallucinations, an issue in which their generated text contains non-existent objects, greatly limiting their reliability and practicality. Current approaches often rely on the model's token likelihoods or other internal information, instruction tuning on additional datasets, or incorporating complex external tools. We first perform empirical analysis on sentence-level LVLM hallucination, finding that CLIP similarity to the image acts as a stronger and more robust indicator of hallucination compared to token likelihoods. Motivated by this, we introduce our CLIP-Guided Decoding (CGD) approach, a straightforward but effective training-free approach to reduce object hallucination at decoding time. CGD uses CLIP to guide the model's decoding process by enhancing visual grounding of generated text with the image. Experiments demonstrate that CGD effectively mitigates object hallucination across multiple LVLM families while preserving the utility of text generation. Codes are available\footnote{\url{https://github.com/d-ailin/CLIP-Guided-Decoding}}.

\end{abstract}

\section{Introduction}

\begin{wrapfigure}{r}{0.4\textwidth} 
    \vspace{-2mm}
    \centering
    \includegraphics[width=\linewidth]{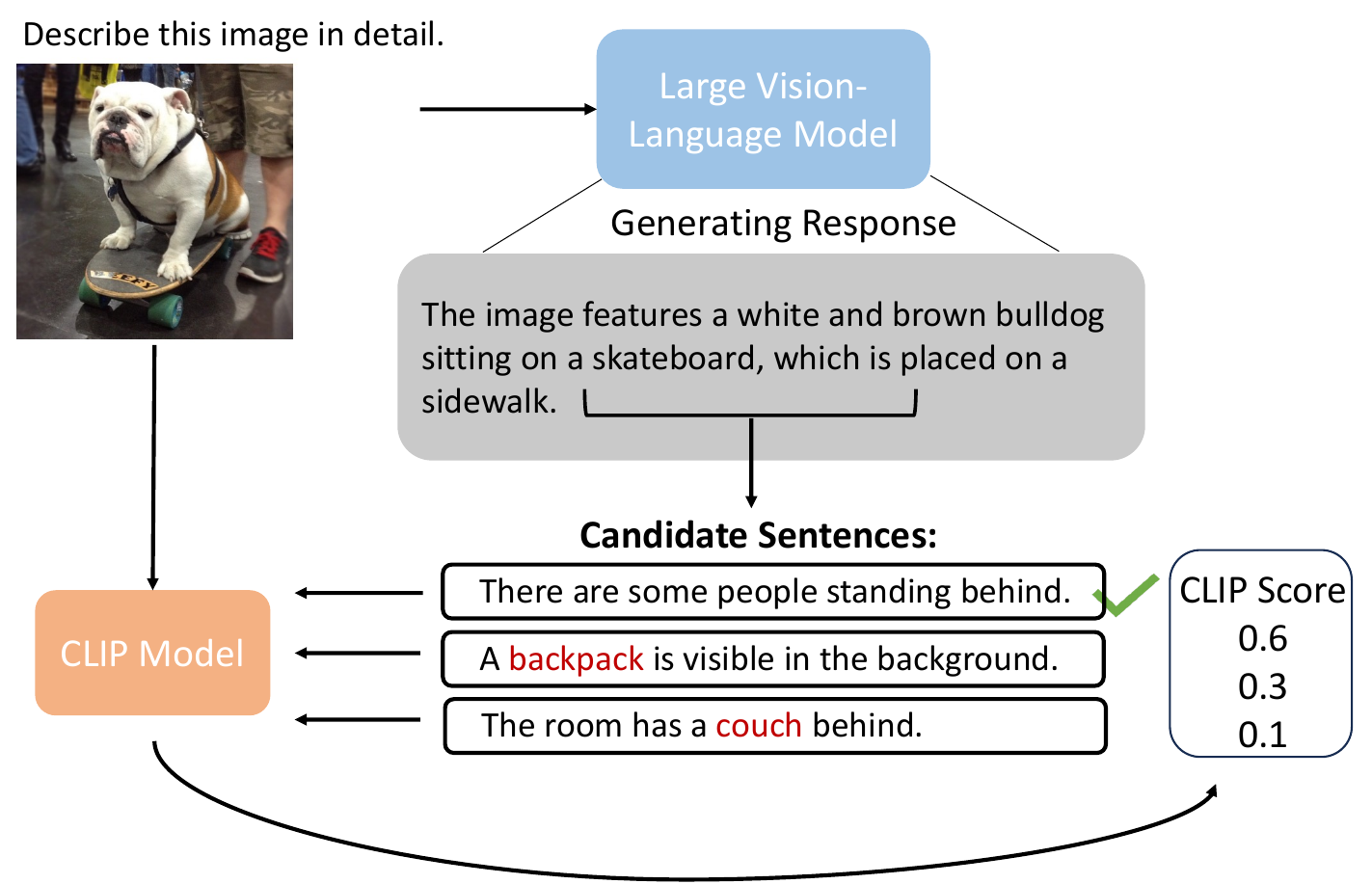}
    \captionsetup{font=small}
    \caption{\small Intuition of our method: candidate sentences with higher CLIP similarity to the image are less likely to be hallucinated, and hence selected during the decoding process. Hallucinated text is colored \textcolor{red}{red}.}
    \label{fig:illustration}
\end{wrapfigure}

Large Vision-Language Models (LVLMs) have shown impressive visual reasoning capabilities, representing an important milestone toward agents that can operate autonomously in our visual world~\citep{achiam2023gpt,liu2023llava,dai2305instructblip,xi2023rise}. However, object hallucination, in which the model produces inaccurate descriptions featuring non-existent objects, greatly limit the model's reliability and practical utility~\citep{rohrbach-etal-2018-object,wang2023evaluation,gunjal2023detecting,zhou2023analyzing}. Hallucinations can easily mislead users, particularly when accompanied by overconfidence~\citep{xiong2023can}. This is especially a concern for safety-critical applications such as robotics~\citep{brohan2023rt} and medical image analysis~\citep{thawkar2023xraygpt}, as well as for human-AI interaction settings where models should behave in a predictable manner that is well aligned with human expectations.

Intuitively, object hallucination can be viewed through a lens of human-AI misalignment. Consider how humans describe images: generally, we mentally `anchor' or compare our descriptions to objects in the image, making hallucination unlikely. In contrast, LVLMs generate text based on token likelihoods without explicitly `anchoring' or comparing to the image in this manner, making hallucination more likely. This gap between how humans and AI operate makes errors made by AI more surprising and unpredictable, which is detrimental in settings involving human-AI interaction.

Recent efforts~\citep{liu2023mitigating, wang2023vigc} have introduced approaches to mitigate hallucinations through instruction tuning. However, these methods come with substantial additional costs, including an annotation budget to acquire extra instruction data~\citep{liu2023mitigating}. Some studies advocate leveraging internal information from models, such as likelihood scores~\citep{zhou2023analyzing} and internal hidden states~\citep{huang2023opera,leng2023mitigating}. Nevertheless, relying solely on a model's internal information may be insufficient~\citep{manakul2023selfcheckgpt} and potentially unreliable, given the known overconfidence issues in neural networks~\citep{kadavath2022language,xiong2023can}. Alternatively, incorporating external tools or knowledge has been suggested~\citep{yin2023woodpecker}, but this approach often involves separate, occasionally intricate modules, accompanied by a heavy engineering burden~\citep{yin2023woodpecker}. 

In contrast with these approaches, we mitigate hallucination by directly comparing the LVLM's generated text to the image to ensure good correspondence between them, improving alignment to how humans perform the task. 
Specifically, as illustrated in Figure \ref{fig:illustration}, we introduce CLIP-Guided Decoding (CGD), a straightforward but effective training-free approach that utilizes CLIP as an external guide to alleviate hallucination during decoding.

CLIP models are widely adopted in image-text evaluation~\citep{hessel2021clipscore} and have primarily been investigated in the context of pairwise comparisons ~\citep{hessel2021clipscore,thrush2022winoground,hsieh2023sugarcrepe}. However, it is still underexplored if CLIP models can identify hallucinations in open-ended texts generated by LVLMs, including those whose vision encoders are themselves CLIP models~\citep{liu2023llava,li2023blip}. 

To study this, we empirically analyze sentence-level object hallucination in LVLMs, finding that CLIP scores are a stronger and more robust indicator of hallucination compared to token likelihoods, especially in the later sentences of each caption. We also observe that hallucination is consistently more likely in later sentences, across multiple LVLMs. Motivated by this, we integrate CLIP models as external guidance into decoding at the sentence level. 
Empirical results demonstrate the success of our method in mitigating hallucination while preserving the utility of text generation.
Interestingly, we observe improvement even when reusing the CLIP models utilized in the LVLMs, indicating the potential overfitting during the fine-tuning phrase in existing LVLMs.
Our contributions can be summarized as follows:
\begin{itemize}
    \item We conduct a comprehensive hallucination analysis across multiple datasets, COCO and NoCaps (Out-of-Domain), utilizing different metrics including likelihood scores and CLIP scores at the sentence level.
    \item We propose CLIP-Guided Decoding (CGD), designed as a lightweight method to facilitate external guidance during decoding at the sentence level, mitigating hallucination in a training-free manner.
    \item We quantitatively assess our method in hallucination evaluation together with generation quality evaluation. The empirical results show our method's effectiveness in reducing hallucination while preserving the overall generation quality.
\end{itemize}

\section{Related Works}
\subsection{Large Vision-Language Models}

Recent developments in Large Vision-Language Models (LVLMs)~\citep{liu2023llava,li2023blip,ye2023mplug} have been significantly powered by the open-sourcing of Large Language Models (LLMs) such as LLaMA~\citep{touvron2023llama} and Vicuna~\citep{chiang2023vicuna}. 
These advancements enable LVLMs to make remarkable strides in understanding and addressing a wide range of vision-language tasks with more integrated capabilities~\citep{yu2023mm,yue2023mmmu}. Most LVLMs share the same two training phases, i.e., pre-trained feature alignment and instruction fine-tuning, to align the vision feature with language features from LLMs and make the model comprehend and follow the instruction~\citep{liu2023llava,dai2305instructblip}. 
While these LVLMs have shown promising improvements in handling more complex and general tasks compared to earlier, smaller vision-language models~\citep{zhou2020unified}, they are still easily suffered from hallucination issues~\citep{li-etal-2023-evaluating,zhou2023analyzing}, especially in open-ended generation contexts~\citep{zhang2023language}.
\vspace{-6mm}
\subsection{Hallucination in VLMs}

While the issue of hallucinations in LLMs has been widely studied in the field of NLP~\citep{ji2023survey}, hallucination mitigations in recent LVLMs are still unexplored. There are recent efforts in mitigating hallucination, including robust instruction tuning~\citep{liu2023mitigating} and using intrinsic information from models, such as likelihood scores or hidden states from the model~\citep{zhou2023analyzing,huang2023opera}. However, only relying on the internal states of models could be potentially unreliable due to the known overconfident issues in neural networks~\citep{kadavath2022language,xiong2023can}. While leveraging external knowledge or models is an alternative way, existing methods~\citep{yin2023woodpecker} often result in substantial engineering complexity.
On the other hand, while vision-language pre-trained models, e.g. CLIP, are widely adopted in evaluation~\citep{hessel2021clipscore} and mainly studied in the pairwise comparison context~\citep{yuksekgonul2022and,thrush2022winoground,hsieh2023sugarcrepe}, the efficacy of CLIP models in detecting hallucination in open-ended generation remains underexplored. Motivated by this, we first conduct hallucination analysis with likelihood-based scores and CLIP scores at the sentence level.

\subsection{Decoding Strategies for Mitigating Hallucinations}
Recent work on reducing hallucinations in LLMs has explored using internal state signals~\citep{chuang2023dola,li2024inference} and contrasting output distributions with different prompts, layers or models~\citep{shi2023trusting,chuang2023dola,li2022contrastive} during decoding. These approaches have also been extended to LVLMs, including focusing on attention patterns~\citep{huang2023opera} or contrasting outputs for varied image inputs~\cite{leng2023mitigating,chen2024halc}. Unlike these methods relying on the internal states or contrasting distributions, our paper introduces using CLIP to guide the generation process towards more visually accurate content in a seamless way.

\section{Preliminaries}
\paragraph{Notations.}
We denote input $\bm{x} = (\bm{x}_{\rm img}, \bm{x}_{\rm text} )$ including an input image $\bm{x}_{\rm img}$ and a text input $\bm{x}_{\rm text}$, e.g. `Describe this image in detail'.
In addition, we represent the generated response $\bm{y}$ in the form of $L$ sequential sentences as 
$ \bm{y} \coloneqq (\bm{s}_1, \dots, \bm{s}_L) $, where a sentence $\bm{s}_i$ is the $i$-th sentence and consists of $l_i$ tokens: $\bm{s}_i \coloneqq ( z_1^{(i)}, \dots, z_{l_i}^{(i)} )$ for $i\in [L]$ and $[L] \coloneqq \{1, \dots, L\}$. Additionally, we define $H(\bm{s}_i) \in \{ 0, 1 \} $ to output the hallucination label given a sentence $\bm{s}_i$. That is, $H(\bm{s}_i) = 1$ if the sentence $\bm{s}_i$ contains an incorrect object
for the image $\bm{x}_{\rm img}$.
\paragraph{Sentence Likelihood.}
Sentence likelihood and length-normalized sentence likelihood are common metrics in conditional language generation~\citep{ranzato2015sequence,wu2016google,adiwardana2020towards,zhao2023calibrating} and are standard baselines in hallucination detection. Since we focus on hallucination detection at the sentence level, we first reformulate the generative process as sentence generation.

Given a Large Vision-Language Model (LVLM) model with parameters $\bm{\theta}$,
an image $\bm{x}_{\rm img}$ and text input $\bm{x}_{\rm text}$, the model generates a response $\bm{y}$ by conditional sentence generation in an auto-regressive manner:
\begin{equation*}
    \log p_{\bm{\theta}} (\bm{y} \given \bm{x} ) 
    = \log p_{\bm \theta}(\bm{s}_1,\dots,\bm{s}_{L} \given \bm{x}) \nonumber
    = \sum_{i=1}^L \log p_{\bm \theta} (\bm{s}_i \given \bm{x}, \bm{s}_{<i} ), \nonumber
\end{equation*}
where $\bm{s}_{<i}$ are the sentences before $i$-th sentence in the response $\bm{y}$.
Further, sentence generation is a process of conditional token generation: 
\begin{equation}
    \log p_{\bm \theta} (\bm{s}_i \given \bm{x}, \bm{s}_{<i} ) = \log \sum_{j=1}^{l_i} p_{\bm \theta}(z_j^{(i)} \given \bm{x}, \bm{s}_{<i}, z_{<j}^{(i)}), \nonumber
\end{equation}
where $z_{<j}^{(i)}$ are the tokens before position j in sentence $\bm{s}_i$.
Though $\log p_{\theta}(\bm{y} \given \bm{x})$ is statistically meaningful, it has been shown to be biased with sequence length, i.e., models tend to overly favor shorter sentences~\citep{wu2016google}. Length-normalized likelihood is an alternative, where the normalization can be either over the response or the sentence:
\begin{align}
    &f_{\bm \theta}(\bm{y}) \coloneqq \frac{1}{\sum_{i=1}^L l_i} \sum_{i=1}^L \log p_{\bm \theta} (\bm{s}_i \given \bm{x}, \bm{s}_{<i} ), \quad
    g_{\bm \theta}(\bm{s}_i) \coloneqq \frac{1}{l_i} \log p_{\bm \theta} (\bm{s}_i \given \bm{x}, \bm{s}_{<i} ),\label{eq:sent_cond_likelihood}
\end{align}
where $f_{\bm \theta}(\bm y)$ represents the length normalized likelihood of a response $\bm{y}$, and $g_{\bm \theta}(\bm{s}_i)$ is the sentence likelihood, the length normalized likelihood of the $i$-th sentence $\bm{s}_i$ conditioning on the previously generated sentences.

\paragraph{CLIP.}
Vision-language models pre-trained with image-text contrastive objectives~\citep{radford2021learning,Zhai_2023_ICCV}, e.g. CLIP, are effective for aligning vision and language domains. Typically, a CLIP model parameterized with $\bm \phi$, consists of image and text feature extractors: $f_{\bm{\phi}_{\rm{img} }}$ and $f_{\bm{\phi}_{\rm{text} }}$.
Given an image $\bm{x}_{\rm img}$ and a sentence $\bm{s}$, we obtain the CLIPScore as cosine similarity between the normalized image feature and text feature: 
\begin{align}
f_{\bm \phi} (\bm{x}_{\rm img}, \bm{s}) \coloneqq \cos({ f_{\bm{\phi}_{\rm{img} }}(\bm{x}_{\rm img}), f_{\bm{\phi}_{\rm{text}}}(\bm{s})} ).
\label{eq:clipscore}
\end{align}
CLIPScore is a quantitative metric to reflect how well a textual description is associated with the image and widely applied in image-text evaluation ~\citep{hessel2021clipscore}.

\begin{figure*}[t]
  \vspace{-10mm}
  \centering
    \captionsetup[subfigure]{skip=2pt} 
  \begin{subfigure}{0.32\textwidth}
    \includegraphics[width=\linewidth]{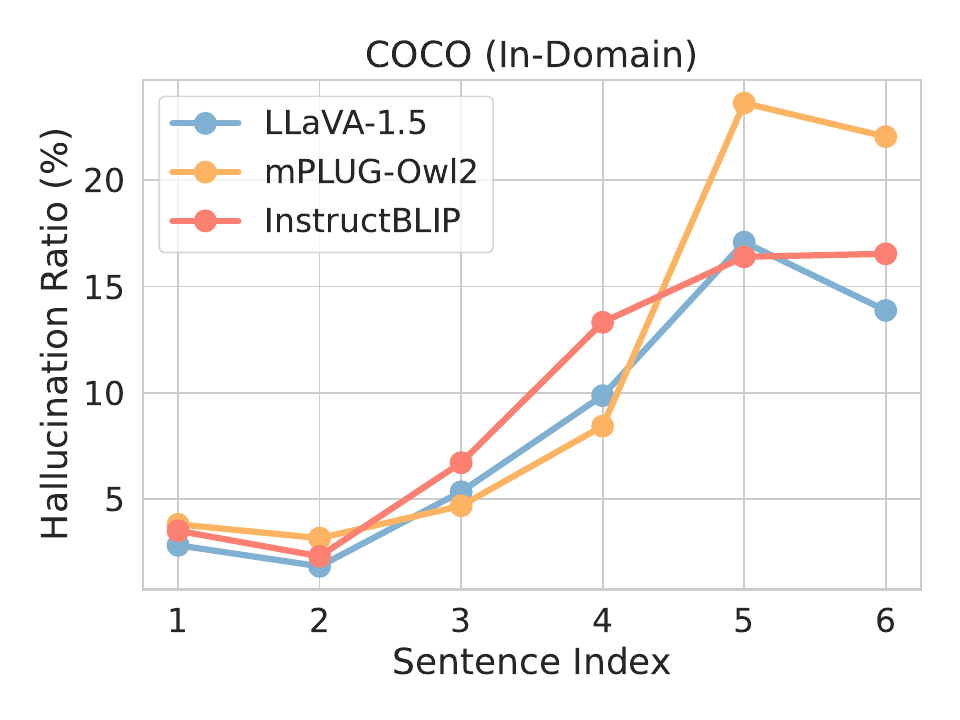}
    \subcaption{}
    \label{fig:hallu_ratios}
  \end{subfigure}%
  \begin{subfigure}{0.34\textwidth}
    \includegraphics[width=\linewidth]{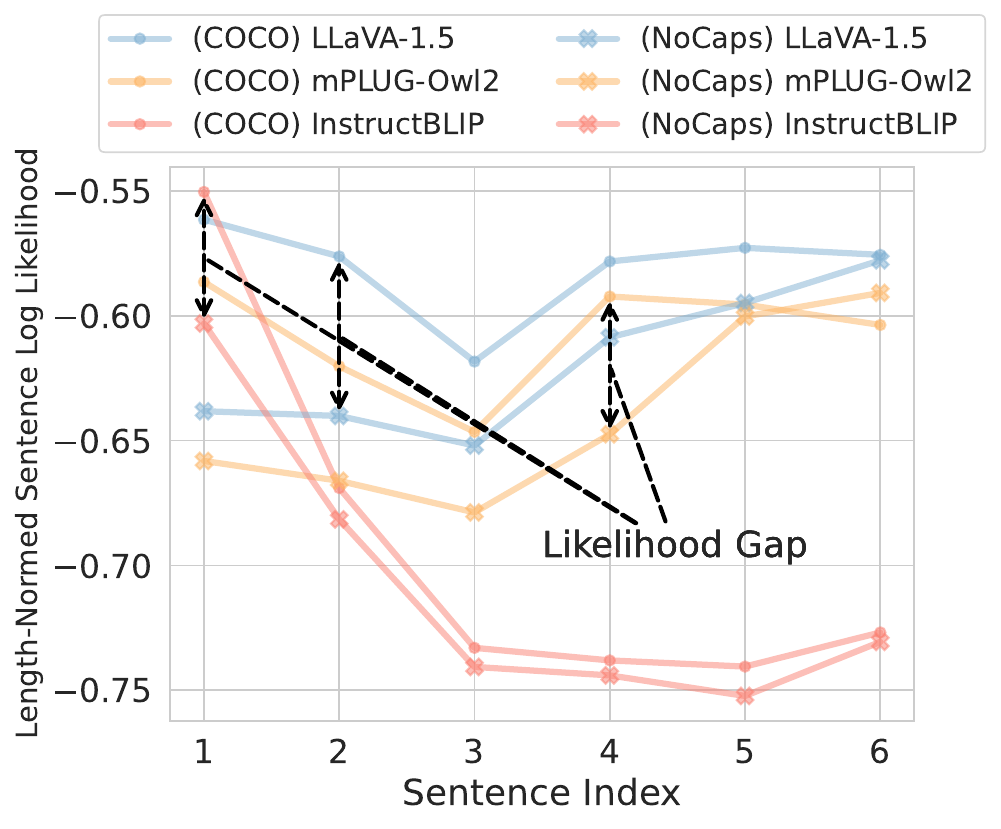}
    \subcaption{}
    \label{fig:likelihood_means}
  \end{subfigure}%
  \begin{subfigure}{0.32\textwidth}
    \includegraphics[width=\linewidth]{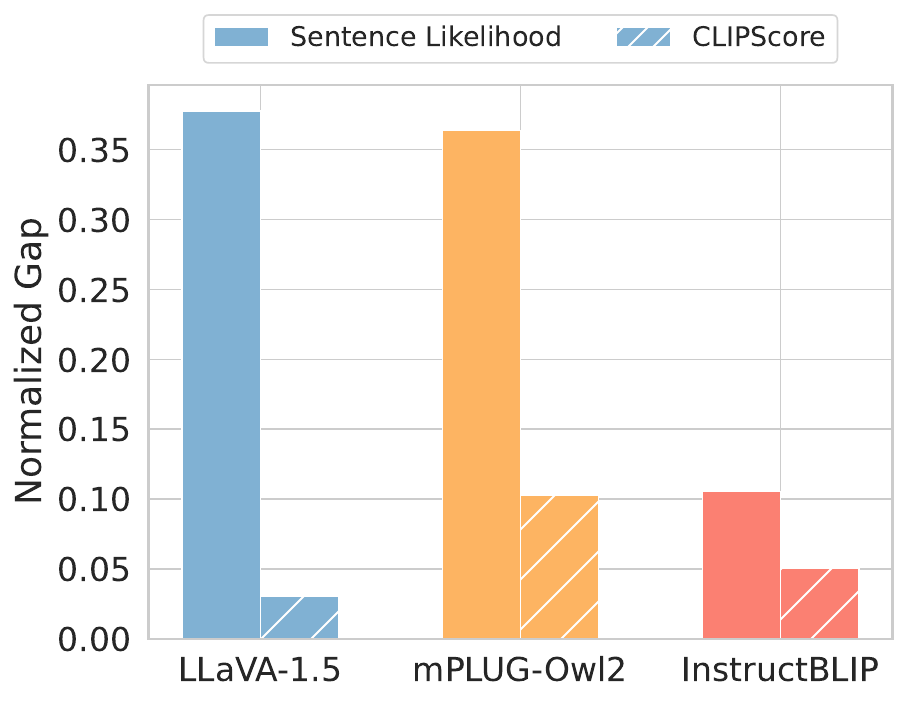}
    \subcaption{}
    \label{fig:gap_comparison}
  \end{subfigure}
  \vspace{-3mm}
  \caption{\small \textbf{(a)} Hallucination ratios in sentences generated by three LVLM models on COCO. The later-generated sentences are more prone to hallucinations. \textbf{(b)} Likelihood gap between COCO and NoCaps over three models implies the potential instability of likelihood-only methods when applied to diverse datasets. \textbf{(c)} Normalized gap between COCO and NoCaps using sentence likelihood and CLIPScore. A smaller gap value indicates more stable values across different datasets. This finding shows that CLIPScore values are more stable across different datasets compared to sentence likelihood. }
  \label{fig:analysis}
\end{figure*}

\subsection{Hallucination Analysis}
In this section, we study how well the sentence likelihood scores and CLIPScore can detect hallucinated sentences generated by LVLMs. 
Note that recent efforts~\citep{zhou2023analyzing} have also studied uncertainty-related metrics for object hallucination, but focus on the token level, whereas our work focuses on sentence-level hallucination. 
As individual tokens lack sufficient semantic meaning, it is challenging to determine if they are hallucinations at the token level without broader context. This issue also necessitates careful token selection in practical applications~\citep{zhou2023analyzing}. In contrast, a sentence, as a more self-contained unit in natural language, offers a clearer and more effective basis for studying hallucinations.

\paragraph{Models.}
We have included three LVLMs: InstructBLIP~\citep{dai2305instructblip}, mPLUG-Owl2~\citep{ye2023mplug} and LLaVA-1.5~\citep{liu2023llava} in the study. We use Greedy Decoding to generate the response with an image and the prompt, `Describe this image in detail'. Note that in our work, we use InstructBLIP (Vicuna-7B), mPLUG-Owl2 (LLaMA-7B) and LLaVA-1.5 (Vicuna-7B) and set the maximum new tokens as $500$ by default. 

\paragraph{Datasets.}
We use images from COCO~\citep{karpathy2015deep} and NoCaps Validation (Out-of-Domain) datasets~\citep{agrawal2019nocaps}.
Specifically, The NoCaps~\citep{agrawal2019nocaps} dataset is proposed to evaluate models trained on the training set of COCO captions to examine how well they generalize to a much larger variety of visual concepts, i.e., unseen objects. NoCaps (Out-of-Domain) set is a subset of NoCaps dataset and includes samples with novel classes, which are unseen in the COCO dataset. Following the procedure in ~\citet{rohrbach-etal-2018-object}, the hallucination label $H(\bm{s}_i)$ for a sentence $\bm{s}_i$ is decided based on whether the sentence contains a non-existent object label when compared with all objects in the ground-truth caption and segmentation labels provided by the datasets. We include $1000$ samples in analysis for each dataset. We include more details in Appendix~\ref{app:experiment}.

\paragraph{Metrics}
To quantify hallucination at sentence level, we define two metrics: hallucination ratio $R(\cdot)$ and first-time hallucination ratio $R_{\rm first}(\cdot)$. For evaluating the detection performance, we use AUROC metric.

Given a dataset $\mathcal{D}$ and the corresponding output $\mathcal{Y}$ generated by a model, we denote $\mathcal{Y}_i \coloneqq \{\bm{y} \in \mathcal{Y} \given |\bm{y}| \geq i \}$ to be the set of responses where each response $\bm{y}$ has a length of at least $i$ sentences; here $|\bm{y}|$ denotes the number of sentences in the response $\bm{y}$. Hence, we define the \emph{hallucination ratio} at $i$-th sentence as the fraction of times the $i$-th sentence is hallucinated, among responses with at least $i$ sentences:
\begin{align}\label{eq:hallu_ratio}
    R(i) &\coloneqq \frac{ | \{ \bm{y} \in \mathcal{Y}_i \given H(\bm{s}_i) = 1 \} | }{ | \mathcal{Y}_i | }.
\end{align}
In addition,  we define the \emph{first-time hallucination ratio} as the fraction of responses where hallucination first occurs at the $i$-th sentence, among responses with at least $i$ sentences:
\begin{align}\label{eq:first_hallu_ratio}
    R_{\rm first}(i) &\coloneqq \frac{ | \{ \bm{y} \in \mathcal{Y}_i \given H(\bm{s}_i) = 1, H(\bm{s}_j) = 0, \forall j < i | }{ | \mathcal{Y}_i | }.
\end{align}
\paragraph{Later Sentences Are More Prone to Hallucinations.}
We investigate the hallucination ratios $R(\cdot)$ across sentences regarding the position of the sentences in the generated description. Figure~\ref{fig:hallu_ratios} shows that the sentences generated in the later part are more prone to hallucination, with surprisingly consistent increasing pattern across multiple LVLMs. This finding echoes previous observations about object positional bias in ~\citet{zhou2023analyzing}, which is at the token level. This bias 
indicates the severity of hallucination as longer descriptions are generated.

Does this positional bias occur solely due to error propagation in sequential generation~\citep{arora2022exposure,zhang2023language}, i.e., early errors inducing later errors? To investigate this, we focus on first-time hallucination ratios $R_{\rm first}(\cdot)$ across different sentence indexes, which removes the effect of error propagation. Interestingly, as shown in Figure~\ref{fig:first-time-hallu-full}, the bias remains evident. This suggests that positional bias is not exclusively a result of error propagation. Instead, it may be partly attributed to diminishing attention to visual inputs as the length of the generated descriptions increases~\citep{zhang2024debiasing}.

\begin{figure*}[t]
  \vspace{-8mm}
  \centering
  \begin{subfigure}{0.32\textwidth}
    \includegraphics[width=\linewidth]{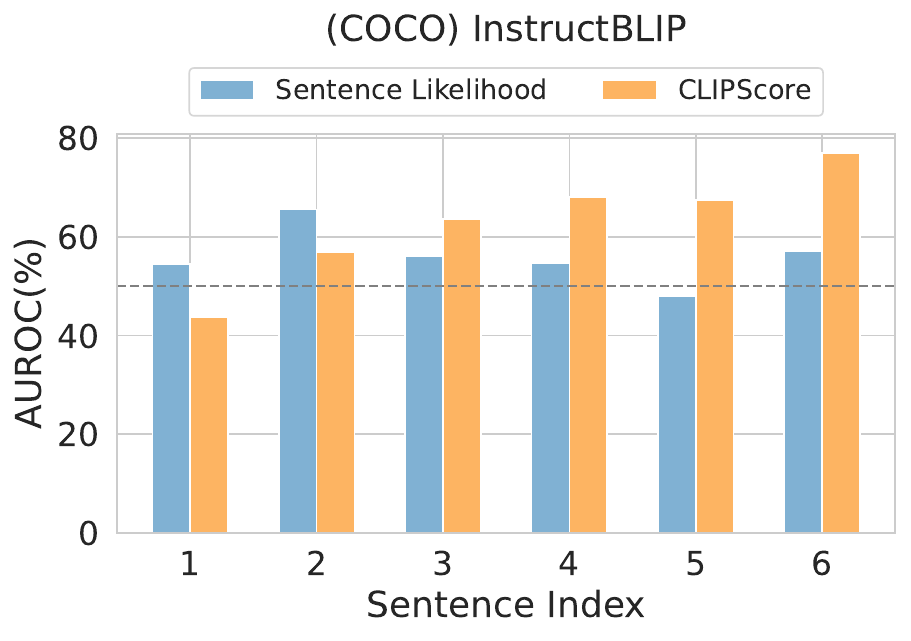}
    \label{fig:}
  \end{subfigure}%
  \begin{subfigure}{0.32\textwidth}
    \includegraphics[width=\linewidth]{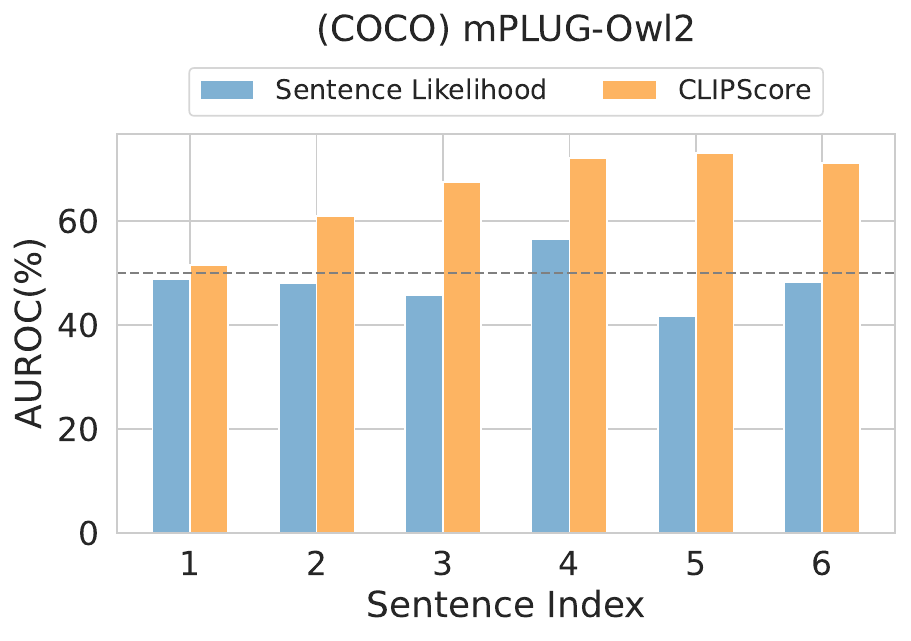}
    \label{fig:}
  \end{subfigure}%
  \begin{subfigure}{0.32\textwidth}
    \includegraphics[width=\linewidth]{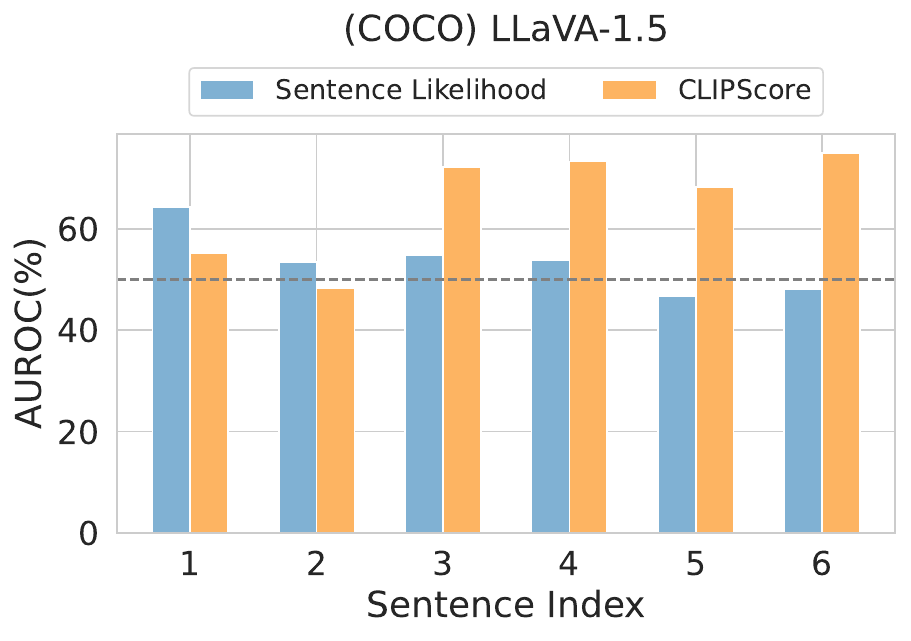}
    \label{fig:}
  \end{subfigure}
  \vspace{-5mm}
  \captionsetup{font=footnotesize}
  \caption{\footnotesize Hallucination detection performance (AUROC) of using Sentence Likelihood $g_{\bm \theta}(\cdot)$ and CLIPScore over COCO dataset with three LVLMs. CLIPScore outperforms sentence likelihood generally, especially for the later sentences. An AUROC of $50$\% represents random guessing; see performance on NoCaps in Figure~\ref{fig:likelihood_vs_clip_full_app} in Appendix.}
  \label{fig:likelihood_vs_clip_coco}
\end{figure*}

\paragraph{Poor Predictive Performance of Sentence Likelihood.}
Next, we assess the performance of hallucination detection, measured by AUROC, using length-normalized sentence likelihood $g_{\bm{\theta}}(\cdot)$ as defined in \Cref{eq:sent_cond_likelihood}. An effective metric should distinguish between hallucinated and non-hallucinated sentences, yielding a higher AUROC value. As depicted in Figure~\ref{fig:likelihood_vs_clip_coco}, the performance of sentence likelihood diminishes in later sentences and exhibits generally weak predictive capabilities in hallucination detection. 

\paragraph{Likelihood Gap of Sentence Likelihood.}
To investigate the stability of likelihood scores over different datasets, in Figure~\ref{fig:likelihood_means} we compare the mean values of sentence likelihood across COCO and NoCaps (Out-of-Domain) datasets. Notably, a discernible gap often exists between the mean likelihood scores over COCO and out-of-domain data. Specifically, likelihood scores obtained in COCO are consistently higher than those in the out-of-domain data, suggesting that models may exhibit more confidence in the data they were trained on, particularly since COCO datasets are commonly utilized for fine-tuning in LVLMs~\citep{liu2023llava,dai2305instructblip}. This observation raises concerns about relying solely on likelihood scores in real-world applications, as these internal metrics from models may be influenced by training data or other model-specific characteristics~\citep{ranzato2015sequence}, making it challenging to generalize them to out-of-domain data.

\paragraph{Can CLIPScore Detect Hallucination?}
Existing studies mainly focus on CLIPScore within a pairwise context, wherein a correct caption is compared with a modified version generated through word manipulation (such as removal, addition, or swapping)~\citep{thrush2022winoground,hsieh2023sugarcrepe}. There is limited exploration of CLIPScore's efficacy in detecting hallucinations within an open-world generation setting.
In contrast to the traditional pairwise approach, our investigation demonstrates that CLIPScore is adept at distinguishing incorrect (hallucinated) sentences from correct ones in a broader context. Illustrated in Figure~\ref{fig:likelihood_vs_clip_coco} and Figure~\ref{fig:likelihood_vs_clip_full_app} in the Appendix,
CLIPScore exhibits notable effectiveness in distinguishing hallucinated and non-hallucinated sentences across different models and datasets. Moreover, as CLIP models function as independent external examiners, they exhibit insensitivity to positional bias, performing well across sentence indexes. We also investigate the stability of scores across COCO and NoCaps datasets. Figure~\ref{fig:gap_comparison} indicates that CLIPScore maintains greater consistency across these datasets when compared to sentence likelihood scores.

\section{CLIP-Guided Decoding}
Given the effectiveness of CLIPScore in detecting hallucination, we propose CLIP-Guided Decoding (CGD) to reduce hallucination by using CLIP as vision-language guidance to perfer visually grounded content during generation. The algorithm involves two parts: \emph{Reliability Scoring}, which designs a scoring function aiming to prioritize candidate responses which are less likely to be hallucinated, and \emph{Guided Sentence Generation}, which generates responses based on this scoring function. We decode in a similar way to beam search, but at the sentence level: this allows us to apply CLIP scoring on full sentences instead of incomplete words or phrases which would be present when decoding at the token level.

\paragraph{Reliability Scoring.}
Given a step $t$ at decoding time and a candidate response $\bold{c}$ containing sequential sentences $(\bm{s}_1, \dots, \bm{s}_t)$, we define the reliability score as follows:
\begin{equation}\label{eq:scoring_func}
    F(\bold{c}) \coloneqq (1-\alpha) f_\theta (\bold{c}) + \alpha  \frac{1}{t} \sum_{i}^{t} f_\phi (\bm{x}_{\rm img}, \bm{s}_i),
\end{equation}
where $\alpha \in [0,1]$ is hyperparameter
to weigh the normalized likelihood $f_\theta (\bold{c})$ from the LVLM model, and the CLIP guidance $\sum_{i}^{t} f_\phi (\bm{x}_{\rm img}, \bm{s}_i)$. 
Mixing between likelihoods and CLIP guidance gives users flexibility to control the strength of CLIP guidance.
When $\alpha = 0$, CLIP guidance has no effect and the scoring function only considers the likelihood scores from the model. $\alpha=1$ indicates the preference for higher CLIPScore. 

\paragraph{Guided Sentence Generation.}
Next, we generate responses, guided by the scoring function. An important goal is to \emph{maintain the generation quality and fluency} of the LVLM by preserving its sentence sampling process. To mitigate hallucinations, we use the reliability score function to prioritize responses that are \emph{well-grounded} to the image.

Specifically, for every step $t$, we maintain a set of candidates with maximum cardinality $N$ as $\mathcal{C}_t \coloneqq \{\bold{c}_1^{t},\bold{c}_2^{t},\dots, \bold{c}_N^{t}\}$, where each candidate $\bold{c}_j^{t}\coloneqq (\bm{s}_{j,1}^{t},\bm{s}_{j,2}^{t},\dots,\bm{s}_{j,t}^{t})$ represents the first $t$ sequences that are generated for $j \in [N]$.

Given a candidate set $\mathcal{C}_t$ in step $t$, the candidate set $\mathcal{C}_{t+1}$ is generated as follows. Firstly, for every $\bold{c}_j^t \in \mathcal{C}_t$,  we independently sample its next sentence $M$ times, following the conditional distribution parameterized by LVLM model parameter $\bm \theta$ (see Algorithm~\ref{alg:clip_guided_decoding}), from which we obtain a new candidate set $\mathcal{C}'_{t +1}$. Then, to avoid the cardinality of the candidate set exponentially increasing, we keep the top $N$ candidates in $\mathcal{C}'_{t +1}$ based on the sentence-level hallucination scoring, i.e.,
\begin{align}\label{eq:topN_candidates}
    \mathcal{C}_{t +1} \coloneqq \underset{\bold{c}\in \mathcal{C}'_{t +1}}{\text{argtop$N$}} \{ F(\bold{c}) \},
\end{align}
where argtop$N$ returns the top $N$ members of the set in the subscript, with the highest values of the function ($F$). 
We repeat the procedure until we reach EOF for all candidates, e.g., reaching the maximum output length. Finally, we use the highest-scoring candidate $\bold{c}^*$ from $\mathcal{C}_{t}$ as output. We summarize CLIP-Guided Decoding in Algorithm~\ref{alg:clip_guided_decoding}.

\section{Experiments}
\vspace{-2mm}
\subsection{Experimental Setup}

\paragraph{Tasks and Datasets.}
Our evaluation tasks include hallucination evaluation and open-ended VQA, where hallucination evaluation is our main focus, and open-ended VQA allows us to gain insight into the method's effect in a wider array of settings.
The hallucination evaluation includes COCO~\citep{lin2014microsoft} (using Karpathy Test split~\citep{karpathy2015deep}), NoCaps (near-domain) and Nocaps (out-of-domain)~\citep{agrawal2019nocaps}, where Nocaps (out-of-domain) contains the out-of-domain data regarding COCO objects. We randomly select $500$ samples from each set for each run and prompt the LVLMs with ``Describe this image in detail''. 
For open-ended VQA, we use the open-sourced multi-modality task benchmark MM-Vet~\citep{yu2023mm} and follow the provided automatic evaluation procedure. See more details in Appendix \ref{app:experiment}.

\vspace{-2mm}
\paragraph{Metrics.}
We follow \citet{zhou2023analyzing} to calculate CHAIR metrics for automatic hallucination evaluation. 
Specifically, we compute CHAIR$_i$ ($C_I$) and CHAIR$_s$ ($C_S$) as follows:
\begin{align*}
    \textsc{CHAIR}_{i} = \frac{| \{\text{hallucinated objects}\} |}{ | \{\text{all objects mentioned}\} | },
    \textsc{CHAIR}_{s} = \frac{| \{\text{captions with hallucinated objects}\} |}{ | \{\text{all captions}\} | }.
\end{align*}
Besides, we include metrics evaluating generation quality, including average number of words per response (Avg. Len), average coverage ratio (Avg. Coverage), which calculates the proportion of correctly identified objects in a generated description relative to the total number of `golden' (i.e., actual or reference) objects present in the image, BLEU~\citep{papineni-etal-2002-bleu}, METEOR~\citep{banerjee-lavie-2005-meteor}, ROUGE-L~\citep{lin-2004-rouge}, CIDEr~\citep{vedantam2015cider}, SPICE~\citep{anderson2016spice} and CLIPScore~\citep{hessel2021clipscore}.

\paragraph{Baselines and CGD Implementation.}
We compare our method with six decoding methods, including three common strategies: greedy decoding, Nucleus sampling and TopK sampling; and DoLa~\citep{chuang2023dola}, VCD~\citep{leng2023mitigating} and OPERA~\citep{huang2023opera}, three recent decoding methods proposed for mitigating hallucinations in LLMs and LVLMs. For our method, we adopt a recent SOTA image-text contrastive model SigLIP~\citep{Zhai_2023_ICCV} as the CLIP-guiding model and set hyperparameters $N = 3$, $M = 3$ and $\alpha=0.99$ by default. We run our experiments on a GeForce RTX 3090 and report average performance over three runs. We include further details in Appendix~\ref{app:experiment}.

\subsection{Experimental Results}

\input{./tables/coco_nocaps}

\paragraph{Hallucination Evaluation on COCO and NoCaps.}
As COCO has been prevalently used in fine-tuning LVLMs~\citep{liu2023llava,dai2305instructblip}, we extend our evaluation to include images from NoCaps dataset that feature object classes less or not presented in COCO. This set includes near-domain and out-of-domain images. Table~\ref{tab:coco_nocaps} displays the empirical results with CHAIR metrics for COCO, near-domain and out-of-domain data. Compared to responses generated in COCO, the average response length of out-of-domain images is shorter, indicating the LVLM models generally output less for their less confident data. As a smaller CHAIR metric reflects a lower fraction of hallucinated objects in the response, our method consistently surpasses other baselines, achieving lower scores in the CHAIR metrics while maintaining response lengths similar to those produced by the baselines.
\vspace{-3mm}
\paragraph{Generation Quality Evaluation on COCO.}
\input{./tables/coco_quality}
An effective decoding strategy is crucial for minimizing hallucinations while maintaining high-quality generation in textual outputs. To assess response quality in LVLMs, we expand our evaluation to include widely recognized caption-related metrics such as CIDEr~\citep{vedantam2015cider} and SPICE~\citep{anderson2016spice}. We calculate these metrics for responses generated by three different models on the COCO dataset. As shown in Table~\ref{tab:coco_quality}, our method performs on par with other approaches across these diverse metrics. This underscores our method's ability to preserve the overall utility of text generation, while reducing hallucination.

\paragraph{Open-Ended VQA Performance on MM-Vet.}

\begin{wrapfigure}[5]{r}{0.3\textwidth}
    \vspace{-12mm}
    \centering
    \includegraphics[width=0.29\textwidth]{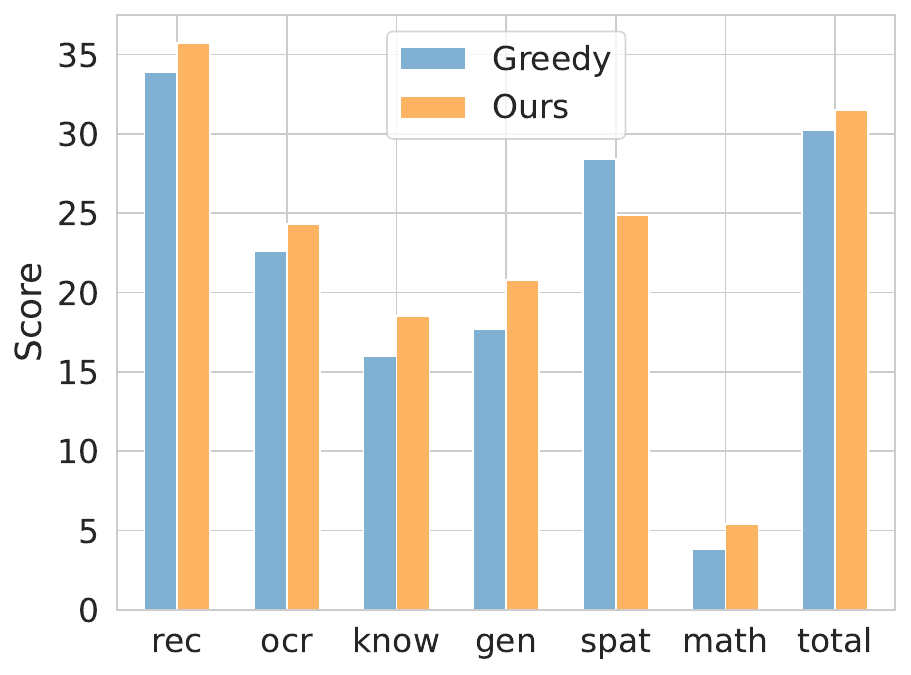}
    \vspace{-4mm}
    \captionsetup{font=scriptsize}
    \caption{\scriptsize Results with LLaVA-1.5 on MM-Vet~\citep{yu2023mm} to evaluate capabilities on multi-modality tasks: recognition (rec), ocr, knowledge (know), language generation (gen), spatial awareness (spat) and math. }
    \label{fig:mmvet}
\end{wrapfigure}

In addition to evaluating hallucination, we conduct a study on our algorithm using the open-ended VQA benchmark, MM-Vet~\citep{yu2023mm}, to gain insights into its effect in a wider array of settings.
Figure~\ref{fig:mmvet} illustrates the empirical comparison between our method and the standard greedy decoding approach. Notably, our method surpasses the baseline in most tasks, such as recognition and OCR. However, it underperforms in spatial awareness tasks, likely attributable to the limitations in relational understanding of CLIP models~\citep{thrush2022winoground,yuksekgonul2022and,hsieh2023sugarcrepe}. Ongoing research seeks to enhance vision-language models through data curation~\citep{fang2023data,xu2023demystifying} or feature fusion techniques~\citep{wang2023sam,tong2024eyes}, which could further augment our method through a better guidance model with enhanced visual grounding capabilities. Case studies are shown in Table~\ref{tab:case_study_mmvet}.

\paragraph{Ablation Study of Hallucination Scoring.}
\begin{wrapfigure}[12]{r}{0.3\textwidth}
    \vspace{-5mm}
    \centering
    \includegraphics[width=0.29\textwidth]{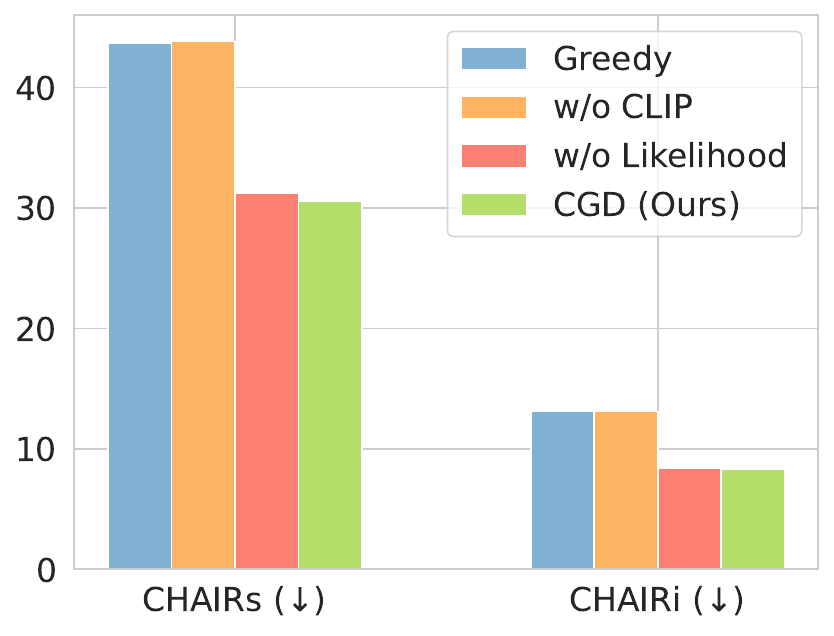}
    \vspace{-4mm}
    \captionsetup{font=scriptsize}
    \caption{\scriptsize Ablation study of hallucination scoring defined in ~\Cref{eq:scoring_func}. The results are averaged across all datasets with LLaVA-1.5. See the numerical results in Table~\ref{tab:ablation_full}.}
    \label{fig:ablation}
\end{wrapfigure}

To further understand the functionality of sentence likelihood and CLIPScore in our hallucination scoring, defined in ~\Cref{eq:scoring_func}, we conduct an ablation study focusing on sentence likelihood and CLIPScore's impact on hallucination mitigation by controlling $\alpha$. Specifically, the algorithm operates without sentence likelihood when $\alpha = 1$ and without CLIP guidance when $\alpha = 0$. From Figure~\ref{fig:ablation} and Table~\ref{tab:ablation_full}, we can see that the performance without CLIP-guidance is close to greedy decoding and sentence likelihood demonstrates relatively limited effectiveness in reducing hallucination. This result confirms the dominant contribution of CLIP guidance in hallucination mitigation.

\begin{wraptable}[11]{r}{0.4\linewidth}
    \centering
    \footnotesize
    \vspace{-4mm}
    \captionsetup{font=footnotesize}
    \caption{\footnotesize Sensitivity and efficiency analysis of maximum candidate number N and sampling times M. Time: inference time (s) per sample.} \label{tab:hyper_nm}
    \vspace{-3mm}
    \resizebox{\linewidth}{!}{
        \begin{tabularx}{1.075\linewidth}{@{}>{\arraybackslash}X@{}}
        \footnotesize
\begin{tabular}{@{}llllrr@{}} 
\toprule
Method & N & M & \(C_S \downarrow\) & \(C_I \downarrow\) & Time \\
\midrule
Greedy & - & - & 44.67 & 13.11 & 3.79 \\
VCD    & - & - & 44.60 & 12.54 & 7.31 \\
OPERA  & - & - & 49.52 & 13.70 & 44.33 \\
CGD    & 1 & 3 & 37.66 &  9.90 & 7.07 \\
       & 1 & 5 & 34.50 & 8.95 & 10.24 \\
        & 3 & 3 & 29.73 & 8.12 & 19.13 \\
        & 3 & 5 & \textbf{28.40} & \textbf{6.78} & 29.81 \\
\bottomrule
\end{tabular}
        \end{tabularx} 
    }
\end{wraptable}

\paragraph{Sensitivity and Efficiency Analysis of Hyperparameters $N$ and $M$.}
\label{sec:hyperparameter}
To examine the sensitivity of our method to variations in maximum candidate number $N$ and sampling times $M$, we carry out the experiments with different settings where $N \in \{1, 3\}$ and $M \in \{3, 5\}$. We omit $M = 1$ from our study as this setting reverts to the ordinary sampling. Table~\ref{tab:hyper_nm} shows that our method outperforms the baseline even with a reduced maximum candidate number and sampling times. Generally, the performance improves with larger $N$ and $M$. This suggests that having a broader range of choices and larger candidate budgets helps the model avoid hallucination. In terms of efficiency, our method is on par with VCD while being more efficient than OPERA, and provides better hallucination mitigation. The inference cost rises with larger N and M, indicating the tradeoff between efficiency and effectiveness.

\begin{wraptable}[7]{r}{0.4\linewidth}
    \centering
    \footnotesize
    \vspace{-4mm}
    \captionsetup{font=footnotesize}
    \caption{\footnotesize Sensitivity analysis of CLIP-guidance model.} \label{tab:hyper_clip}
    \vspace{-3mm}
    \resizebox{\linewidth}{!}{
        \begin{tabularx}{1.075\linewidth}{@{}>{\arraybackslash}X@{}}
        \footnotesize
\begin{tabular}{@{}llr@{}}
\toprule
Method & \(C_S \downarrow\) & \(C_I \downarrow\) \\
\midrule
Greedy & 44.67 & 13.11 \\
w/ SigLIP ViT-SO-14@384px & 29.73 &  8.12 \\
w/ OpenAI ViT-L-14@336px & 32.80 &  8.78 \\
\bottomrule
\end{tabular}
        \end{tabularx} 
    }
\end{wraptable}

\paragraph{Effect of CLIP Guiding Model.}
As most of the LVLMs have adopted CLIP models from OpenAI as vision encoders, it is worthwhile to investigate whether utilizing the same CLIP models as guiding models in our method can still enhance hallucination mitigation. For a fair comparison, we maintain all the settings but substitute the guidance model with CLIP ViT/L-14 with $336$px resolution image input~\citep{radford2021learning}, which is the vision encoder in LLaVA-1.5. Table~\ref{tab:hyper_clip} shows that our method still surpasses the baseline method even when using this vision encoder. This finding suggests that the existing fine-tuning processes for vision-language alignment in LVLMs~\citep{liu2023llava,dai2305instructblip} might, to some extent, compromise the original vision capabilities. 
Our method is a way to recalibrate the alignment between vision and language by directly examining the outputs from LVLMs.

\vspace{-1mm}
\section{Conclusion}
In this study, we focus on object hallucination analysis and mitigation for open-ended generation in Large Vision-Language Models (LVLMs). We reveal that there exists severe hallucination in later sentences, and CLIPScore is a stronger and more robust indicator of hallucination than token likelihood. Motivated by this, we integrate CLIP as external guidance in a training-free approach during decoding. Our approach effectively mitigates hallucination while preserving generation quality.

\bibliography{main}
\bibliographystyle{colm2024_conference}

\input{appendix}

\end{document}

%% file: math_commands.tex
\usepackage{amsmath,amsfonts,bm}

\def\eqref#1{equation~\ref{#1}}

\def\1{\bm{1}}

\DeclareMathAlphabet{\mathsfit}{\encodingdefault}{\sfdefault}{m}{sl}
\SetMathAlphabet{\mathsfit}{bold}{\encodingdefault}{\sfdefault}{bx}{n}

\DeclareMathOperator*{\argmax}{arg\,max}

%% file: tables/coco_nocaps.tex
\begin{table*}[t]
\centering
 \vspace{-5mm}
 \vspace{-2mm}
 \scriptsize
 \setlength{\tabcolsep}{4pt} 
 \renewcommand{\arraystretch}{0.8} 
\begin{adjustbox}{max width=\textwidth}
\begin{tabular}{c|l|ccc|ccc|ccc}
\toprule
   &  & \multicolumn{3}{c|} {InstructBLIP}  &\multicolumn{3}{c|} {mPLUG-Owl2}  & \multicolumn{3}{c} {LLaVA-1.5}       \\
  & &$C_S$ $\downarrow$ & $C_I$ $\downarrow$ & Avg. Len 
  &$C_S$ $\downarrow$ & $C_I$ $\downarrow$ & Avg. Len
  &$C_S$ $\downarrow$ & $C_I$ $\downarrow$ & Avg. Len \\
   \midrule
\multirow{7}{*}{\shortstack{COCO}} & 
Greedy & 57.9 & 17.1 & 102.7 & 52.7 & 16.0 & 89.4  & 44.7 & 13.1 & 80.1 \\
& Nucleus  &56.1 & 17.0 & 98.3 &  51.9 & 15.6 & 89.0 &  43.3 & 13.1 & 80.1 \\
& TopK &  55.8 & 16.9 & 97.3 & 53.1 & 15.9 & 89.2 &  44.9 & 13.2 & 79.7 \\
& DoLa & 55.6 & 17.0 & 97.1 & 52.6 & 15.2 & 88.8 & 46.6 & 13.6 & 80.3 \\
& VCD & 63.2 & 19.5 & 92.5 & 51.4 & 16.0 & 89.6 & 44.6 & 12.5 & 85.8 \\
& OPERA & 51.5 & 15.6 & 85.8 & 48.5 & 16.1 & 86.1 &  49.5 &	13.7 & 85.7 \\
\cmidrule{2-11}
& \textbf{\ours\ (ours)} & \textbf{42.7} & \textbf{10.9} & 99.6 & \textbf{35.7} &\textbf{ 8.6} & 85.1 &  \textbf{29.7} & \textbf{8.1} & 76.7 \\
\midrule

\multirow{7}{*}{\shortstack{NoCaps\\(Near-Domain)}} & 
Greedy & 55.7 & 15.4 & 102.9 & 39.5 & 10.1 & 77.7  &  46.1 & 12.2 & 80.0 \\
& Nucleus  & 54.1 & 14.7 & 98.9 &  40.9 & 10.6 & 78.0 &  45.4 & 11.9 & 79.5 \\
& TopK &  53.2 & 14.1 & 98.6 & 41.6  & 10.9 & 78.1 &  46.6 & 12.1 & 79.8 \\
& DoLa & 54.5 & 14.0 & 97.9 & 41.0 & 10.8 & 77.9 & 44.8 & 12.1 & 79.5 \\
& VCD & 60.4 & 16.8 & 94.4 & 40.6 & 10.7 & 79.4 & 46.6 & 11.3 & 86.1 \\
& OPERA & 46.2 & 12.8 & 86.8 & 38.5 & 10.3 & 79.5 & 44.3 & 11.6 & 84.1 \\
\cmidrule{2-11}
& \textbf{\ours\ (ours)} & \textbf{42.6} & \textbf{12.3} & 100.6 & \textbf{29.0} & \textbf{6.9} & 72.8 &  \textbf{33.3} & \textbf{7.9} & 75.9 \\
\midrule
\multirow{7}{*}{\shortstack{NoCaps\\(Out-of-Domain)}} & 
Greedy & 51.3 & 17.3 & 98.4 & 36.4 & 12.4 & 68.0  &  40.2 & 14.2 & 69.1\\
& Nucleus  & 48.8 & 16.3 & 92.8  & 36.8 & 12.1 & 68.8  &  40.7 & 14.3 & 69.4 \\
& TopK &  50.3 & 17.0 & 92.2 & 37.1 & 12.3 & 68.2 &  39.9 & 13.5 & 69.5 \\
& DoLa & 48.0 & 16.3 & 92.3 & 35.9 & 13.2 & 68.3 & 43.5 & 14.4 & 71.1 \\
& VCD & 56.9 & 19.5 & 87.8 & 37.9 &	13.9 & 71.2 & 38.7 & 12.4 & 74.7 \\
& OPERA & 43.9 & 15.5 & 80.8 & 33.9 & 12.3 & 69.0 & 38.1 & 12.5 & 71.7 \\
\cmidrule{2-11}
& \textbf{\ours\ (ours)} & \textbf{42.0} & \textbf{11.6} & 94.1 & \textbf{26.6} & \textbf{8.4} & 65.7 & \textbf{28.8} & \textbf{9.0} & 66.3 \\
\bottomrule
\end{tabular}
\end{adjustbox}
\captionsetup{font=small}
 \caption{\small Hallucination evaluation on COCO Karpathy Test Split~\citep{karpathy2015deep} and NoCaps Validation Set~\citep{agrawal2019nocaps}. Please see the case study in Table~\ref{tab:case_study_coco} in Appendix.}
\label{tab:coco_nocaps}
\end{table*}

%% file: tables/coco_quality.tex
\begin{table*}[t]
    \centering
    \setlength\tabcolsep{2.5pt}
    \footnotesize
    \setlength{\tabcolsep}{4pt} 
    \renewcommand{\arraystretch}{0.8} 
    \begin{adjustbox}{max width=\textwidth}
    \begin{tabular}{crcccccccccc}
    \toprule
    Model &  & 



     \makecell{Avg.\\Length} & \makecell{Avg.\\Coverage} & \makecell{BLEU-1} & \makecell{BLEU-2} & \makecell{BLEU-3} & \makecell{BLEU-4} & \makecell{METEOR} & \makecell{ROUGE-L} & \makecell{SPICE} & \makecell{CLIPS} \\

    \midrule
    \multirow{7}*{InstructBLIP}
    & Greedy & 102.65 & 81.10  & 15.85 & 10.97 & 7.02 & 4.41 & 17.12 & 16.99 & 17.69 & 27.06 \\
    & Nucleus & 98.30 & 80.61  & 16.38 & 11.29 & 7.21 & 4.53 & 17.45 & 17.46 & 18.17 & 27.05 \\
    & TopK & 97.26 & 80.04  & 16.46 & 11.26 & 7.13 & 4.43 & 17.36 & 17.48 & 18.00 & 27.03 \\
    & DoLa & 97.11 & 80.11  & 16.54 & 11.34 & 7.22 & 4.50 & 17.47 & 17.61 & 18.11 & 27.07 \\
    & VCD & 92.50 & 79.71 & 17.70 & 10.74 & 6.52 & 3.99 & 17.20 & 17.05 & 17.40 & 26.17 \\
    & OPERA & 85.84 & 79.34 & 18.30 & 12.19 & 7.72 & 4.82 & 18.42 & 19.69 & 18.89 & 27.32 \\
    & \ours & 99.64 & 79.44 & 16.38 & 11.30 & 7.19 & 4.50 & 17.44 & 17.49 & 18.46 & 28.24 \\

    \midrule
    \multirow{7}*{mPLUG-Owl2}
    & Greedy  & 89.35 & 81.28 & 18.08 & 12.59 & 8.28 & 5.39 & 18.79 & 19.28 & 19.18 & 27.08 \\
    & Nucleus & 88.98 & 81.68 & 18.12 & 12.60 & 8.26 & 5.36 & 18.78 & 19.38 & 19.30 & 27.05 \\
    & TopK & 89.15 & 81.36 & 18.09 & 12.51 & 8.18 & 5.29 & 18.76 & 19.38 & 19.22 & 27.06 \\
    & DoLa & 88.76 & 81.30 & 18.14 & 12.58 & 8.20 & 5.29 & 18.81 & 19.46 & 19.20 & 27.07 \\
    & VCD & 89.58 & 78.25 & 17.47 & 11.90 & 7.58 & 4.75 & 18.03 & 18.71 & 17.76 & 26.98 \\
    & OPERA & 86.06 & 76.58 & 17.91 & 12.14 & 7.88 & 5.04 & 18.41 & 19.75 & 18.40 & 27.08 \\ 
    & \ours & 85.06 & 80.17 & 19.07 & 13.42 & 8.88 & 5.77 & 19.37 & 20.35 & 20.21 & 28.21 \\

    \midrule
    \multirow{7}*{LLaVA-1.5}
    & Greedy & 80.05 & 81.30 & 18.91 & 12.73 & 8.07 & 5.07 & 18.69 & 20.23 & 18.49 & 26.94 \\
    & Nucleus & 80.14 & 81.19 & 18.88 & 12.77 & 8.10 & 5.12 & 18.73 & 20.25 & 18.50 & 26.93 \\
    & TopK & 79.71 & 81.39 & 18.90 & 12.69 & 8.05 & 5.07 & 18.66 & 20.20 & 18.39 & 26.92 \\
    & DoLa & 80.29 & 80.88 & 18.74 & 12.46 & 7.83 & 4.88 & 18.55 & 19.98 & 17.98 & 26.81 \\
    & VCD & 85.80 & 81.96 & 18.25 & 12.44 & 7.96 & 5.03 & 18.57 & 19.62 & 18.48 & 27.18\\
    & OPERA & 85.71 & 82.92 & 18.11 & 12.15 & 7.73 & 4.88 & 18.48 & 19.61 & 18.18 & 27.16\\
    & \ours & 76.66 & 79.03 & 19.85 & 13.46 & 8.60 & 5.43 & 19.34 & 21.13 & 19.37 & 27.96 \\

    \bottomrule
    \end{tabular}
    \end{adjustbox}
    \captionsetup{font=small}
    \caption{\small Generation quality evaluation on COCO Karpathy Test Split~\citep{karpathy2015deep}.}
    \label{tab:coco_quality}

\end{table*}

%% file: appendix.tex
\newpage
\appendix
\onecolumn
\section{Analysis Detail}

\begin{figure*}[h]
  \centering
  \begin{subfigure}{0.45\textwidth}
    \includegraphics[width=\linewidth]{figures/hallu_ratio_coco.pdf}
    \caption{}
    \label{fig:}
  \end{subfigure}%
  \begin{subfigure}{0.45\textwidth}
    \includegraphics[width=\linewidth]{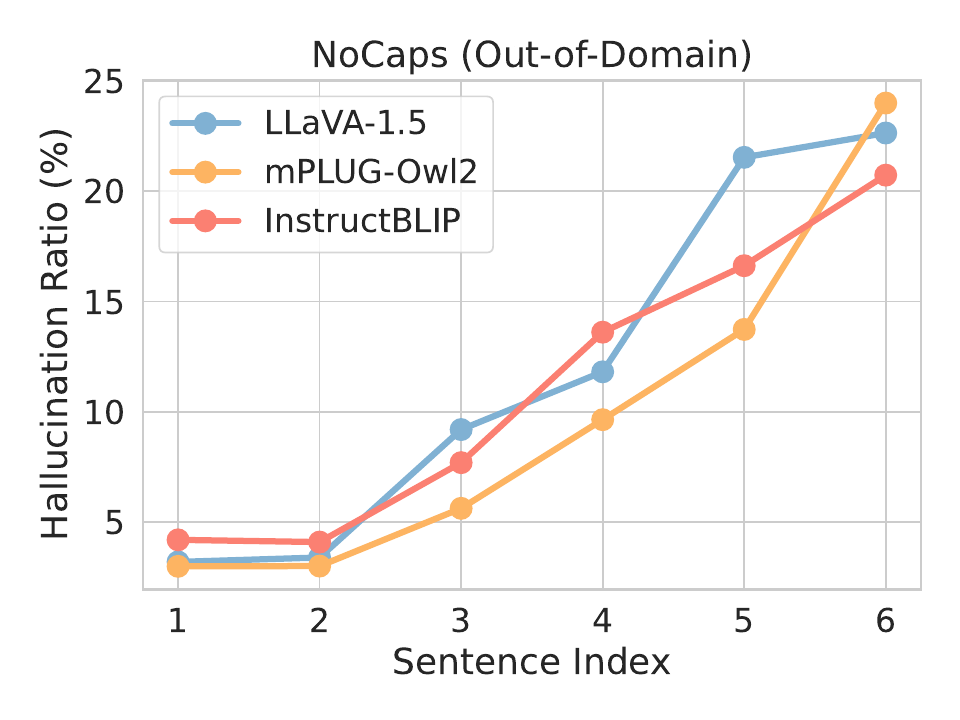}
    \caption{}
    \label{fig:}
  \end{subfigure}%

  \caption{Hallucination ratios across different sentence indexes on both COCO and NoCaps (Out-of-Domain) datasets.}
  \label{fig:hallu-full}
\end{figure*}

\begin{figure*}[h]
  \centering
  \begin{subfigure}{0.45\textwidth}
    \includegraphics[width=\linewidth]{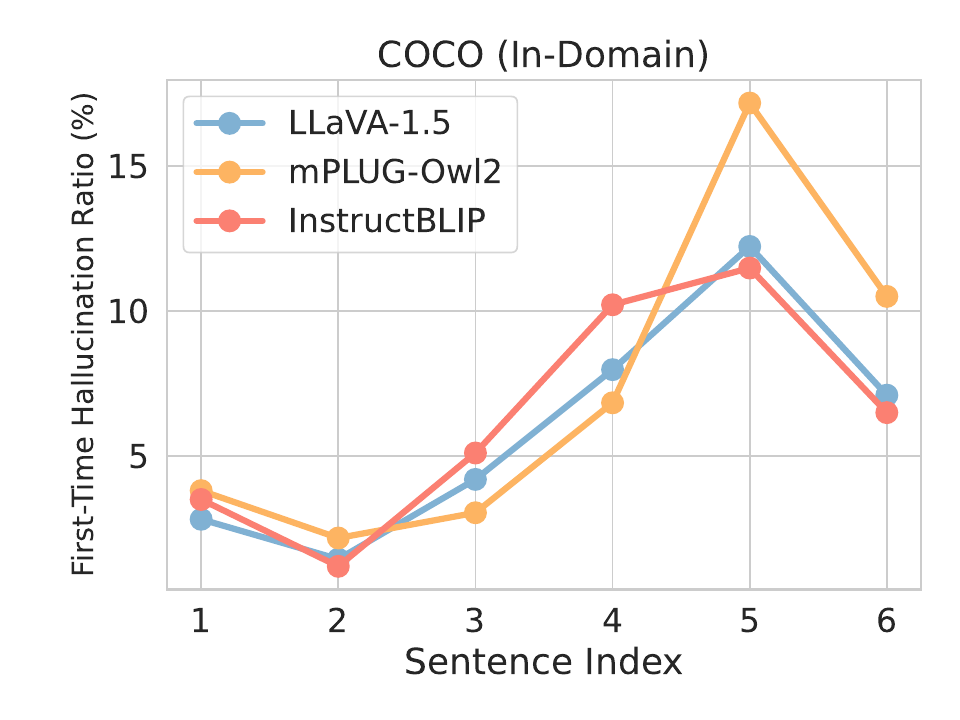}
    \caption{}
    \label{fig:}
  \end{subfigure}%
  \begin{subfigure}{0.45\textwidth}
    \includegraphics[width=\linewidth]{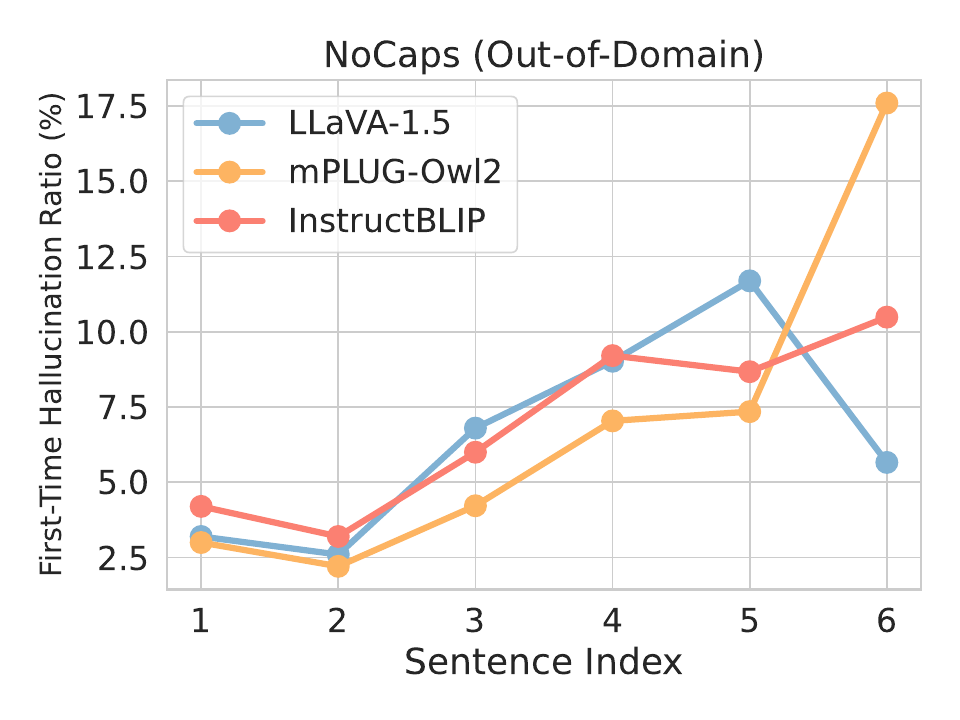}
    \caption{}
    \label{fig:}
  \end{subfigure}%

  \caption{First-time hallucination ratios across different sentence indexes on both COCO and NoCaps (Out-of-Domain) datasets.}
  \label{fig:first-time-hallu-full}
\end{figure*}

\begin{figure*}[h]
  \centering
  \begin{subfigure}{0.32\textwidth}
    \includegraphics[width=\linewidth]{figures/likehood_vs_CLIP_aurocs_COCO_InstructBLIP.pdf}
    \label{fig:}
  \end{subfigure}%
  \begin{subfigure}{0.32\textwidth}
    \includegraphics[width=\linewidth]{figures/likehood_vs_CLIP_aurocs_COCO_mPLUG-Owl2.pdf}
    \label{fig:}
  \end{subfigure}%
  \begin{subfigure}{0.32\textwidth}
    \includegraphics[width=\linewidth]{figures/likehood_vs_CLIP_aurocs_COCO_LLaVA-1.5.pdf}
    \label{fig:}
  \end{subfigure}
  
    \medskip
  \begin{subfigure}{0.32\textwidth}
    \includegraphics[width=\linewidth]{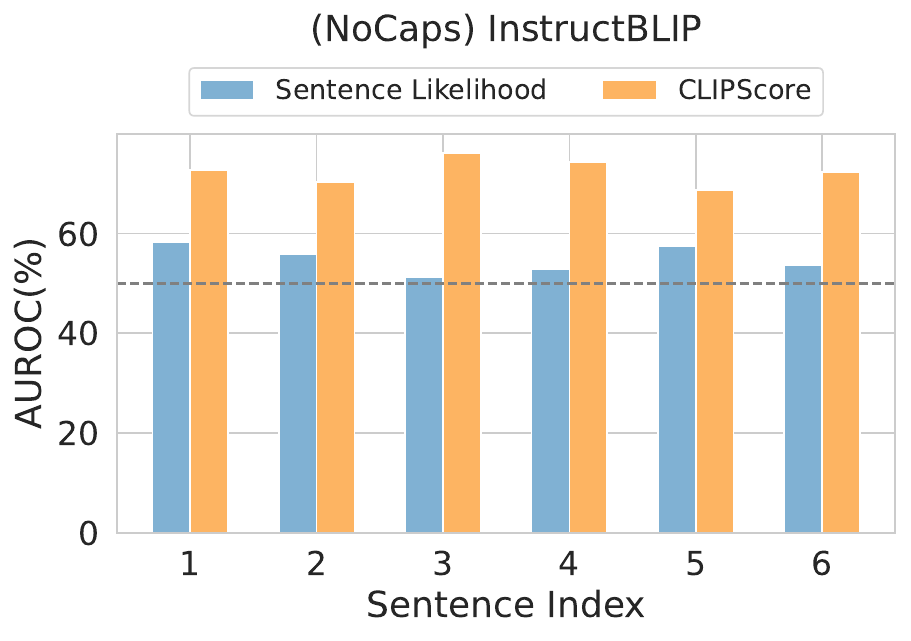}
    \label{fig:}
  \end{subfigure}%
  \begin{subfigure}{0.32\textwidth}
    \includegraphics[width=\linewidth]{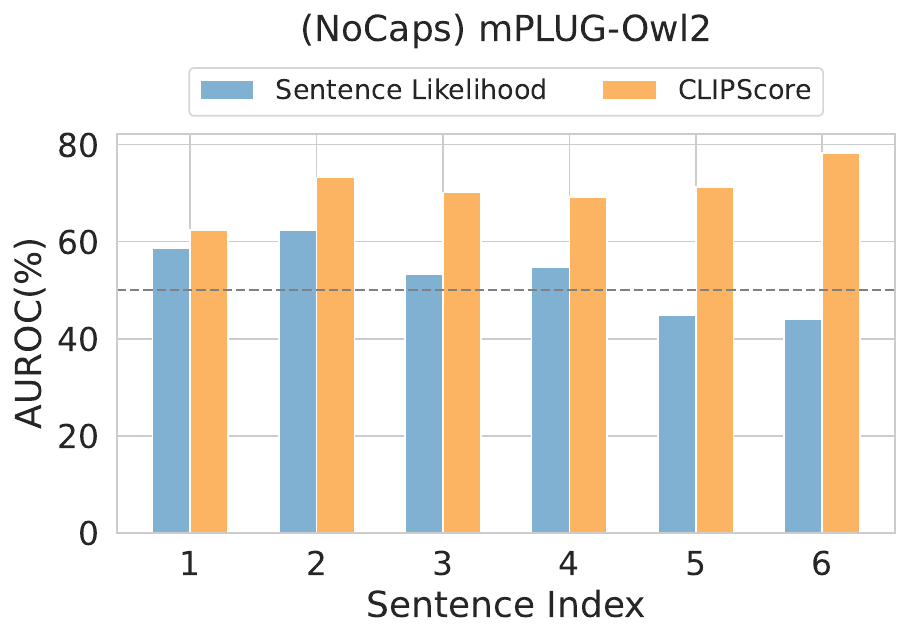}
    \label{fig:}
  \end{subfigure}%
  \begin{subfigure}{0.32\textwidth}
    \includegraphics[width=\linewidth]{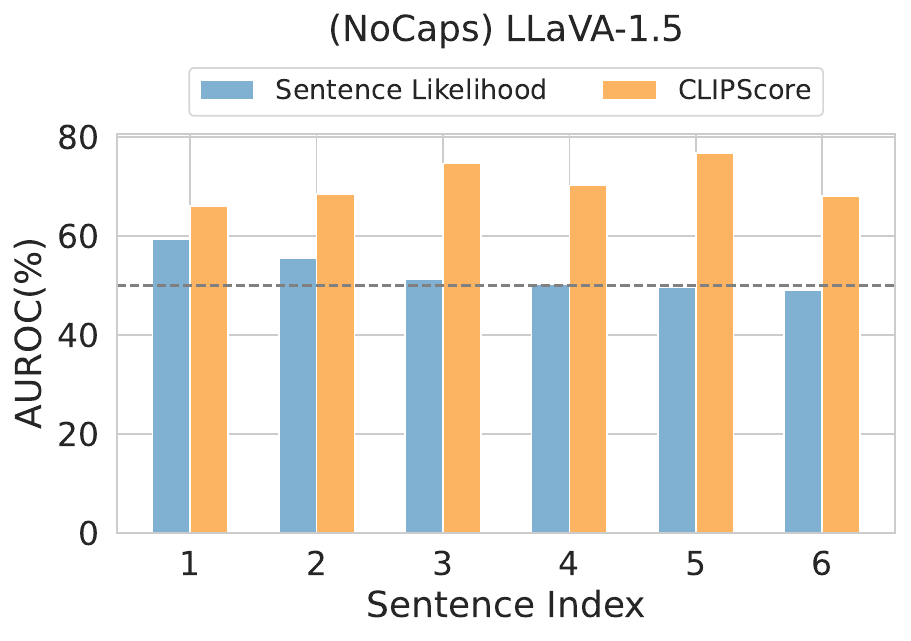}
    \label{fig:}
  \end{subfigure}%

  \caption{Hallucination detection performance (AUROC) of using Sentence Likelihood $g_{\bm \theta}(\cdot)$ and CLIPScore over COCO and NoCap (out-of-domain) datasets with three LVLMs. CLIPScore outperforms sentence likelihood generally, especially for the later sentences. An AUROC of $50$\% represents random guessing. }
  \label{fig:likelihood_vs_clip_full_app}
\end{figure*}

\section{Algorithm}
\label{app:algo}

\begin{algorithm}[H]
  \footnotesize
  \caption{CLIP-Guided Decoding}\label{alg:clip_guided_decoding}
  \begin{algorithmic}[1]
    \State \textbf{Input:} input containing an image and textual prompt $\bm{x} \coloneqq (\bm{x}_{\rm img}, \bm{x}_{\rm text}) $, LVLM parameterized by $\bm \theta$, CLIP model parameterized by $\bm \phi$, maximum cardinality $N$, sampling size $M$, weight hyperparameter $\alpha$
    \State \textbf{Output:} Response output $\bold{c}^*$

    \State $t \coloneqq 0$
    \State $\mathcal{C}_0 \coloneqq \{\langle\text{start token}\rangle\}$.
    
    
    \While{not EOF}
      \State $\mathcal{C}'_{t+1} \coloneqq \varnothing$
      
      \For{$\bold{c}_i^{t} \in \mathcal{C}_{t}$}
        \Repeat
          \State $\bm{s} \sim \mathbb{P}_{\theta} (\bm{s}_{t+1} \given \bm{x}, \bm{s}_{i,1}^t,\ldots, \bm{s}_{i,t}^t)$
          \State $\mathcal{C}'_{t+1} \coloneqq \, \mathcal{C}'_{t+1} \bigcup\, \{ (\bm{s}_{i,1}^t,\ldots, \bm{s}_{i,t}^t, \bm{s}) \} $
        \Until{M times}
      \EndFor
      
      \State $\mathcal{C}_{t+1} \coloneqq $ Top $N$ candidates in $\mathcal{C}'_{t+1}$ \Comment{\Cref{eq:topN_candidates}}
      \State $t \coloneqq t + 1$
    \EndWhile
    
    \State $\bold{c}^{*} \coloneqq \argmax_{\bold{c} \in \mathcal{C}_{t}} F(\bold{c})$
    \Comment{\Cref{eq:scoring_func}}
    \State \textbf{Return} $\bold{c}^{*}$
  \end{algorithmic}
\end{algorithm}

\section{Experimental Setup} \label{app:experiment}
\subsection{Models}
We select three representative LVLM models for evaluation, including InstructBLIP~\cite{dai2305instructblip}, mPLUG-Owl2~\cite{ye2023mplug} and LLaVA-1.5~\cite{liu2023llava}. InstructBLIP adopts a Q-former to bridge the features between the vision and text modalities, using 32 tokens as image token embeddings. mPLUG-Owl2 and LLaVA-1.5 use linear projection layers to align the vision and text modalities, with 256 or 576 tokens as image representations. Generally, the image encoders of these LVLM models are pre-trained models trained with image-text contrastive objective, e.g., CLIP~\cite{radford2021learning} and EVA-CLIP~\cite{fang2023eva}. Note that we use InstructBLIP (Vicuna-7B), mPLUG-Owl2 (LLaMA-7B) and LLaVA-1.5 (Vicuna-7B) and set the maximum new tokens as $500$ by default for all methods.

\subsection{Datasets}

\paragraph{COCO.}
The COCO dataset~\cite{lin2014microsoft} is a comprehensive dataset used for image recognition, segmentation and captioning. It contains over $300,000$ images spanning over $80$ object categories, each with detailed annotations. Given the detailed and high-quality annotation, most recent LVLMs use samples from COCO for vision-language alignment and instruction tuning~\cite{liu2023llava,li2023blip}.
Specifically, we use samples from COCO Karpathy Split~\cite{karpathy2015deep} in our experiments. 

\paragraph{NoCaps.}
The NoCaps dataset~\cite{agrawal2019nocaps} is proposed for evaluating the models trained with COCO with images less or not seen in the COCO object categories. There are $4,500$ images in the validation set and $10,600$ images in the test set. Images are taken from the Open Images V4~\cite{krasin2017openimages} dataset, which spans $600$ object classes. Due to the unavailability of ground truth captions of the test set, we use the validation set of NoCaps.

\paragraph{MM-Vet.}
MM-Vet~\cite{yu2023mm} is an evaluation benchmark\footnote{\url{https://github.com/yuweihao/MM-Vet}} that examines large multimodal models on complicated multimodal tasks.  It defines $6$ core vision-language capabilities including recognition, OCR, knowledge, language generation, spatial awareness and math. It examines the $16$ integrations of interest derived from the combination of these $6$ capabilities. In total, this benchmark contains $200$ images and $218$ questions, all paired with their respective ground truths, which are human annotated or gathered from the internet.
Specifically, the benchmark has gathered $187$ images from various online sources, $10$ 
high-quality images from VCR~\cite{zellers2019recognition} and $3$ images from ChestX-ray14~\cite{wang2017chestx}. The benchmark has also labeled the capacities required for 
answering each question correctly.

\subsection{Metrics.}
\paragraph{Hallucination Evaluation.}
We follow the guidelines in~\cite{zhou2023analyzing} to calculate CHAIR metrics for automatic hallucination evaluation. More precisely, CHAIR quantifies the degree of object hallucination in a given image description by computing the ratio of all objects mentioned in the description but not present in the ground-truth label set. It comprises two assessment dimensions: $\text{CHAIR}_{s}$ ($C_S$) calculated at the sentence-level, and $\text{CHAIR}_{i}$ ($C_I$) calculated at the instance-level. 
Specifically, CHAIR computes CHAIR$_i$ and CHAIR$_s$ as follows:
\begin{align*}
    \textsc{CHAIR}_{i} &= \frac{| \{\text{hallucinated objects}\} |}{ | \{\text{all objects mentioned}\} | }, \\
    \textsc{CHAIR}_{s}& = \frac{| \{\text{captions with hallucinated objects}\} |}{ | \{\text{all captions}\} | }.
\end{align*}
Following~\citet{rohrbach-etal-2018-object,zhou2023analyzing}, we restrict the objects in $80$ COCO object classes for the COCO dataset. For NoCaps, we set a similar setting and map the fine-grained classes defined in NoCaps to coarse-grained categories based on the hierarchical object relationships in Open Images\footnote{\scriptsize \url{https://storage.googleapis.com/openimages/web/download_v7.html\#df-classes-hierarchy}} to improve the effectiveness of CHAIR metrics. Specifically, we only add the super-categories defined in Open Images to our final object list. Eventually, we construct a list of $90$ coarse-grained object categories from $600$ fine-grained object classes.

\paragraph{Generation Quality Evaluation.}\label{app:metric_quality}
To reflect the quality of the generated description, we include several metrics: average word number in a response (Avg. Length) and average coverage ratio (Avg. Coverage), which calculates the proportion of correctly identified objects in a generated description relative to the total number of 'golden' (i.e., actual or reference) objects present in the image. We also include
other caption-related evaluation metrics including BLEU~\cite{papineni-etal-2002-bleu}, METEOR~\cite{banerjee-lavie-2005-meteor}, ROUGE-L~\cite{lin-2004-rouge}, CIDEr~\cite{vedantam2015cider}, SPICE~\cite{anderson2016spice} and CLIPScore~\cite{hessel2021clipscore}:

\begin{itemize}
    \item \textbf{BLUE} BLUE (Bilingual Evaluation Understudy~\cite{papineni-etal-2002-bleu}) is a metric for evaluating the quality of machine-generated translations by comparing them to one or more reference translations. The BLEU score is based on precision of $n$-grams, which are contiguous sequences of $n$ words. 
    \item \textbf{METEOR} METEOR (Metric for Evaluation of Translation with Explicit ORdering~\cite{banerjee-lavie-2005-meteor}) is designed to evaluate the quality of machine-generated text (translations or captions) by comparing it to one or more human reference texts. Specifically, METEOR computes a harmonic mean (F-mean) of precision and recall, incorporating both the quality of the generated text and its similarity to reference text. 
    \item \textbf{ROUGE-L} ROUGE-L (Recall-Oriented Understudy for Gisting Evaluation - Longest Common Subsequence~\cite{lin-2004-rouge}) is a common metric in evluating text summarization tasks. It is proposed to measure the quality of a machine-generated summary by comparing it to one or more reference summaries.  
    \item \textbf{CIDEr} CIDEr (Consensus-based Image Description Evaluation~\cite{vedantam2015cider}) is designed for image captioning evaluation by measuring the quality of generated image captions by comparing them to human-generated reference captions. The score utilizes the weighted combination of $n$-gram similarity scores in both the generated and reference captions. Specfically, it gives more weight to higher-order $n$-grams to encourage diversity in generated captions.
    \item \textbf{SPICE} SPICE (Semantic Propositional Image Caption Evaluation~\cite{anderson2016spice}) metric is an evaluation measure used for image captioning. Instead of focusing on $n$-grams or surface-level text similarity, SPICE parses both the generated and reference captions into semantic propositions, including objects, attributes, relationships, and actions present in the image. It aims to capture the accuracy of generated captions with an aspect of semantic meanings.
    \item \textbf{CLIPScore} CLIPScore~\cite{radford2021learning,hessel2021clipscore} is a metric based on CLIP (Contrastive Language-Image Pretraining~\cite{radford2021learning}) to measure how well a given textual description is associated with the image by computing the cosine similarity based on the normalized features through CLIP models. 
\end{itemize}

\paragraph{Open-Ended VQA Benchmark Evaluation on MM-Vet.}
For evaluation of open-ended generation, due to the high flexibility and free-form of answers, MM-Vet has proposed an LLM-based evaluator for open-ended outputs. Specifically, they provide a template with the question, ground-truth answer and prediction (e.g., output from an LVLM) in a few-shot prompt and prompt the LLMs to provide a soft grading score from $0$ to $1$, where a high score indicates a more accurate prediction. The total scores are computed by the average scores obtained for all questions. The score regarding each capability is the average score obtained for the questions that have been annotated for the requirement of the specific capability.

\subsection{Baselines.}
\label{app:baselines}
We compare our method with six decoding methods, including three common strategies: greedy decoding, Nucleus sampling and TopK sampling; and DoLa~\citep{chuang2023dola}, VCD~\citep{leng2023mitigating} and OPERA~\citep{huang2023opera}, three recent decoding methods proposed for mitigating hallucinations in LLMs and LVLMs. All methods share a temperature of $0.2$. We set $\text{top-p} = 0.7$ for Nucleus sampling and $\text{top-k} = 5$ for TopK sampling. For DoLa, we use the default setting from the paper~\citep{chuang2023dola}: layers with layer index $0,2,4,6,8,10,12,14$ as the candidate premature layers and $32$ index layer as the mature layer. For VCD, we use default setting: amplification factor as $1$, $\beta = 0.1$ and noise step number as $500$. For OPERA, we set scale factor as $50$, threshold as $15$, number of candidates per beam as $5$ and $1$ as the weight of penalty term in decoding.

\subsection{Implementation Detail.}
We adopt a recent SOTA image-text contrastive model SigLIP~\citep{Zhai_2023_ICCV} as the CLIP-guiding model by default. We also include the results of using variant CLIP models in the sensitivity
study. Following the mainstream sampling method to generate diverse responses~\citep{wang2023evaluation}, we use TopK sampling during generation, with the same setting as the TopK sampling baseline. We run our experiments on a GeForce RTX 3090 and report average performance over three runs. For hyperparameters, we set $N = 3$, $M = 3$ and $\alpha=0.99$ in our algorithm by default. For the ablation and sensitivity study, we focus on the settings with LLaVA-1.5.


\section{Experimental Results}

\subsection{Ablation Study}
Table~\ref{tab:ablation_full} displays the numeric results of the ablation study with different components in the scoring function 
as defined in~\Cref{eq:scoring_func}.
\input{tables/hyper_alpha_full}

\subsection{Case Study}
\label{app:case_study}

\paragraph{Hallucination Case Study.}
We have included several cases of hallucination mitigation on the COCO dataset, as shown in Table~\ref{tab:case_study_coco}. This table shows examples by comparing the original generated description and generated description by our method given the image in the leftmost column. The hallucinated part in the description has been highlighted in red.
\input{tables/case_study}

\paragraph{VQA Case Study.}
We have demonstrated some examples from the MM-Vet dataset in Table~\ref{tab:case_study_mmvet}. We have inlcuded some samples with each row containing the image, the corresponding question, the originally generated response, a response based on our method and the provided ground-truth answer. The score under each response reflects the accuracy of the response to the question. We also show the required capabilities for answering each question in the ``Answer" Column.

\input{tables/case_study_mmvet}

%% file: tables/hyper_alpha_full.tex
\begin{table}[h]
    \centering
    \scriptsize
    \begin{tabular}{lccc|ccc|ccc}
    \toprule
        Method & \multicolumn{3}{c|} {COCO}  &\multicolumn{3}{c|} {NoCaps (Near-Domain)}  & \multicolumn{3}{c} {NoCaps (Out-Domain)}\\
      &$C_S$ $\downarrow$ & $C_I$ $\downarrow$ & Avg. Len 
      &$C_S$ $\downarrow$ & $C_I$ $\downarrow$ & Avg. Len
      &$C_S$ $\downarrow$ & $C_I$ $\downarrow$ & Avg. Len\\    
    \midrule
    Greedy & 44.67  & 13.11 & 80.05 & 46.13 & 12.19 & 79.99 & 40.20 & 14.21 & 69.07 \\
    \midrule
    \ours\ (Ours) & 29.73 & 8.12 & 76.66 & \textbf{33.27} & \textbf{7.86} & 75.86 & \textbf{28.80} & \textbf{8.99} & 66.34 \\ 
    \small{w/o Sentence Likelihood}  & \textbf{29.45} & \textbf{7.37} & 76.12 & 34.33 & 8.38 & 75.47 & 29.93 &	9.47 & 65.88 \\
    \small{w/o CLIP-Guidance} & 45.07 & 13.29 & 81.60 & 44.73 &	11.55 &	81.25 & 41.67 & 14.46 & 70.88 \\

    \bottomrule
    \end{tabular}    
    \caption{Ablation study on scoring components in~\Cref{eq:scoring_func}.}
    \label{tab:ablation_full}
\end{table}

%% file: tables/case_study.tex
\newcolumntype{b}{X}
\newcolumntype{s}{>{\hsize=.6\hsize}X}
\newcolumntype{m}{>{\hsize=1.2\hsize}X}

\begin{table*}[h]
    \small
    \centering
    \begin{tabularx}{\textwidth}{s|m|m}
        \toprule
        \makecell{\textbf{Image}} & \makecell{\textbf{Original}} & \makecell{\textbf{Ours}} \\
        \midrule
        \multirow[c]{3}{*}{\includegraphics[width=0.45\linewidth]{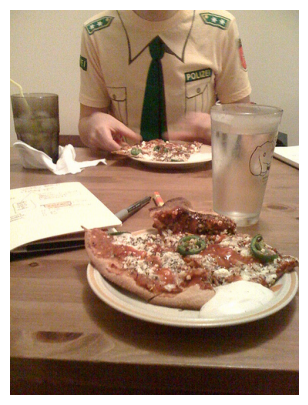}}
        & The image depicts a man sitting at a dining table, enjoying a meal of pizza. He is wearing a police uniform, indicating his occupation. There are two pizzas on the table, one closer to the man and the other further away. The table is set with a cup, \textcolor{red}{a fork, and a knife}. A book is also present on the table, possibly for the man to read while he eats. The scene is set in a cozy environment, with a \textcolor{red}{couch} visible in the background. 
        & The image features a man wearing a police uniform sitting at a dining table with a plate of pizza in front of him. The table is set with two cups, one on the left side and the other on the right side. There are also two slices of pizza on the table, one in front of the man and the other on the right side. The man appears to be enjoying his meal, and the scene seems to be a casual dining experience.
        \\
        \midrule
        \multirow[c]{3}{*}[-0.5cm]{\includegraphics[width=0.6\linewidth]{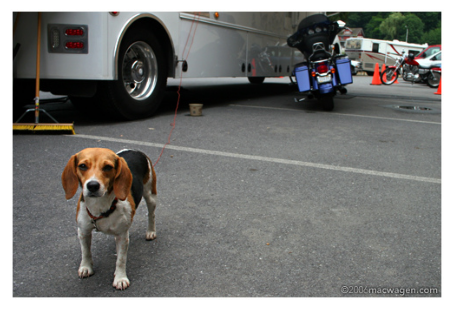}}
        & The image features a small dog standing on a paved road, wearing a collar. The dog is positioned near the center of the scene. A motorcycle is parked on the right side of the dog, and a bus is visible in the background, occupying a large portion of the image. \textcolor{red}{There are two people in the scene. One person is standing near the left edge of the image, while the other person is located on the right side, closer to the motorcycle}.
        & The image features a small brown and white dog standing on a paved surface, possibly a parking lot. The dog is on a leash, and it appears to be looking at the camera. In the background, there is a bus parked, and a motorcycle is also visible in the scene. The dog is positioned near the center of the image, with the bus and motorcycle occupying the right side of the frame.
        \\
        \midrule
        \multirow[c]{3}{*}{\includegraphics[width=0.6\linewidth]{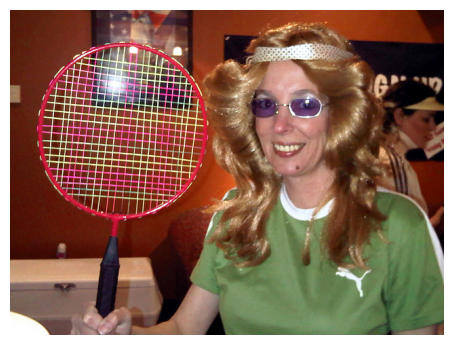}}
        & The image features a woman wearing a blonde wig and holding a tennis racket. She is smiling and appears to be enjoying herself. The woman is standing in a room with \textcolor{red}{a couch} in the background. \textcolor{red}{There is a TV mounted on the wall, and a cup can be seen placed on a surface nearby.} The overall atmosphere of the scene is light-hearted and fun.
        & The image features a woman wearing a blonde wig and a green shirt, holding a tennis racket. She appears to be dressed as a tennis player, possibly for a costume party or a themed event. The woman is smiling and posing with the tennis racket, showcasing her enthusiasm for the sport.
        \\
        \bottomrule
    \end{tabularx}
    \caption{Examples of COCO with LLaVA-1.5. Comparison between the text generated by Original (TopK Sampling with $K=5$) and our methods. The hallucinated parts are highlighted in \textcolor{red}{red}.}
    \label{tab:case_study_coco}
\end{table*}

%% file: tables/case_study_mmvet.tex
\newcolumntype{b}{X}
\newcolumntype{s}{>{\hsize=.5\hsize}X}
\newcolumntype{m}{>{\hsize=1\hsize}X}

\begin{table*}[h]
    \small
    \centering
    \begin{tabularx}{\textwidth}{s|s|m|m|m}
        \toprule
        \makecell{\textbf{Image}} & \makecell{\textbf{Question}} & \makecell{\textbf{Original}} & \makecell{\textbf{Ours}} & \makecell{\textbf{Answer}} \\
        \midrule
        \multirow[c]{3}{*}{\includegraphics[width=0.55\linewidth]{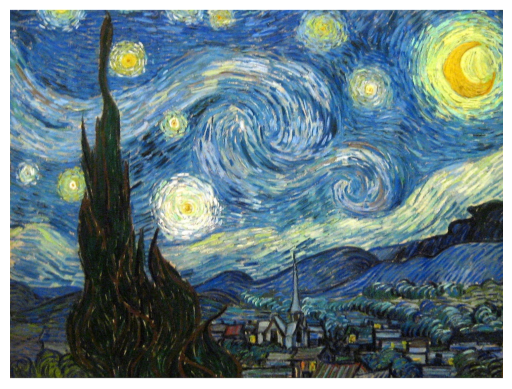}}
        &  Can you give a short introduction to this painting?
        &  This is a painting of a night scene with a city in the foreground and a mountain in the background. The painting is a representation of the night sky with the stars and the moon. The city is depicted with buildings and the mountain is shown with a church on top. The painting is a beautiful representation of the night sky and the cityscape.
        \newline\newline \textbf{(score: 0.2)}
        &  This painting is a famous representation of the night sky by the artist Vincent Van Gogh. The painting is a depiction of the night sky with the stars visible, and it is set against a backdrop of a town. The painting is a combination of the artist's imagination and the night sky, creating a unique and captivating piece of art.
        \newline\newline \textbf{(score: 0.6)}
        &  The Starry Night is an oil-on-canvas painting by the Dutch Post-Impressionist painter Vincent van Gogh. Painted in June 1889, it depicts the view from the east-facing window of his asylum room at Saint-Rémy-de-Provence, just before sunrise, with the addition of an imaginary village. It has been in the permanent collection of the Museum of Modern Art in New York City since 1941, acquired through the Lillie P. Bliss Bequest. Widely regarded as Van Gogh's magnum opus, The Starry Night is one of the most recognizable paintings in Western art.
        \newline\newline \textbf{Required capabilities: recognition, knowledge, language generation}\\
        \midrule
        \multirow[c]{3}{*}{\includegraphics[width=0.55\linewidth]{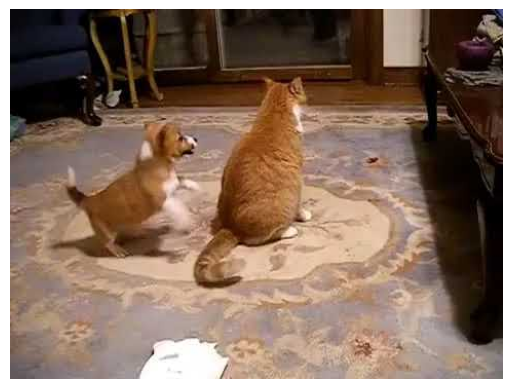}}
        &  Which animal in the picture is larger, the cat or the dog?
        &  The cat is larger than the dog.
        \newline\newline \textbf{(score: 1.0)}
        &  The dog is larger than the cat in the picture.
        \newline\newline \textbf{(score: 0.0)}
        &  cat
        \newline\newline \textbf{Required capabilities: recognition, spatial awareness}\\
        \midrule
        \multirow[c]{3}{*}{\includegraphics[width=0.55\linewidth]{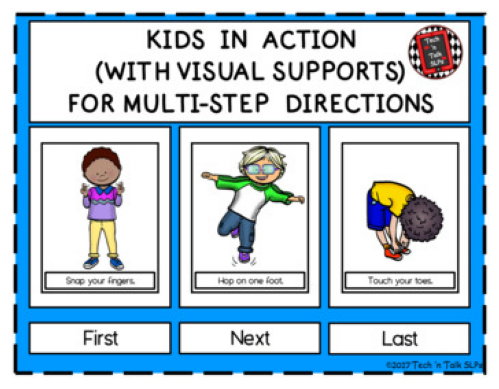}}
        &  What should kids do after snap fingers?
        \newline\newline\newline
        &  After snap fingers, kids should then touch their toes.
        \newline\newline \textbf{(score: 0.0)}
        &   After snap fingers, kids should hop on one foot.
        \newline\newline \textbf{(score: 1.0)}
        &   hop on one foot
        \newline\newline \textbf{Required capabilities: OCR, spatial awareness}\\
        \midrule
        \multirow[c]{3}{*}{\includegraphics[width=0.5\linewidth]{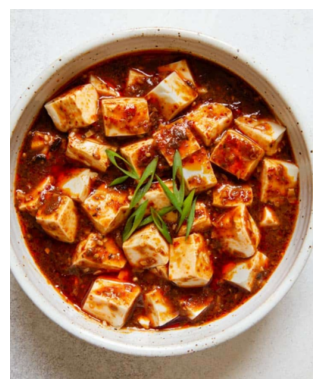}}
        &   What is the name of this dish?
        \newline\newline\newline
        &  The name of this dish is "Tofu Soup" or "Tofu Stew."
        \newline\newline \textbf{(score: 0.0)}
        &   The dish is called "Tofu in a spicy sauce".
        \newline\newline \textbf{(score: 0.4)}
        &    mapo tofu
        \newline\newline \textbf{Required capabilities: recognition}\\
        
        \bottomrule
    \end{tabularx}
    \caption{Examples from MM-Vet~\cite{yu2023mm} based on LLaVA-1.5.}
    \label{tab:case_study_mmvet}

\end{table*}